\documentclass[11pt]{article}

\usepackage[]{acl}

\usepackage{times}
\usepackage{latexsym}

\usepackage[T1]{fontenc}
\usepackage[utf8]{inputenc}

\usepackage{microtype}

\usepackage{inconsolata}

\usepackage{graphicx}
\usepackage{amsmath}
\usepackage{amssymb}
\usepackage{booktabs}
\usepackage{cleveref}
\usepackage{tabularx}
\usepackage{multirow}
\usepackage{float}
\usepackage{dblfloatfix}

\usepackage{fvextra}
\usepackage{setspace}
\usepackage{xcolor}
\definecolor{commentgray}{gray}{0.5}
\definecolor{commentbg}{rgb}{0.9,0.95,1}

\newcommand{\cmtc}[1]{\hfill\colorbox{commentbg}{#1}\hfill{}}

\DefineVerbatimEnvironment{promptblock}{Verbatim}{
  breaklines=true,
  breakanywhere=true,
  fontsize=\fontsize{9}{9},
  frame=single,
  framesep=1mm,
  breaksymbolleft={},
  commandchars=\\\{\}
}

\DefineVerbatimEnvironment{promptblocksml}{Verbatim}{
  breaklines=true,
  breakanywhere=true,
  fontsize=\fontsize{8}{8},
  frame=single,
  framesep=1mm,
  breaksymbolleft={},
  commandchars=\\\{\}
}

\PassOptionsToPackage{dvipsnames,table}{xcolor} %
\usepackage{xcolor}
\usepackage{colortbl}
\definecolor{GainNeg}{RGB}{248,215,218}    %
\definecolor{GainSmall}{RGB}{255,229,204}  %
\definecolor{GainMed}{RGB} {233,245,196}    %
\definecolor{GainLarge}{RGB}{198,239,206}  %
\definecolor{GainLargeText}{RGB}{0,97,0}   %

\newcommand{\gainbox}[2]{%
  \begingroup
  \setlength{\fboxsep}{0.6pt}%
  \colorbox{#1}{\strut #2}%
  \endgroup
}

\newcommand{\gain}[1]{%
  \begingroup
  \edef\val{#1}%
  \ifdim \val pt < 0pt
    \gainbox{GainNeg}{\val}%
  \else
    \ifdim \val pt < 2pt
      \gainbox{GainSmall}{+\val}%
    \else
      \ifdim \val pt < 4pt
        \gainbox{GainMed}{+\val}%
      \else
        \gainbox{GainLarge}{\textcolor{GainLargeText}{\textbf{+\val}}}%
      \fi
    \fi
  \fi
  \endgroup
}

\newcommand{\tierneg}{\gainbox{GainNeg}{\textbf{negative}}}
\newcommand{\tiersmall}{\gainbox{GainSmall}{small}}
\newcommand{\tiermed}{\gainbox{GainMed}{medium}}
\newcommand{\tierlarge}{\gainbox{GainLarge}{\textcolor{GainLargeText}{\textbf{large}}}}

\title{What Does LLM Refinement Actually Improve?\\A Systematic Study on Document-Level Literary Translation}

\author{Shaomu Tan$^{1,3,}$\thanks{ \hspace{1mm} Work done while interning at Amazon.} \qquad Dawei Zhu$^{3}$ \qquad Ke Tran$^{3}$ \qquad Michael Denkowski$^{3}$  \\ \textbf{Sony Trenous$^{3}$ \qquad Bill Byrne$^{2,3}$ \qquad Leonardo Ribeiro$^{3}$ \qquad Felix Hieber$^{3}$} \\[0.9em]
  $^{1}$University of Amsterdam \qquad $^{2}$University of Cambridge \qquad $^{3}$Amazon AGI \\[0.2em]
  \texttt{s.tan@uva.nl} \qquad \{daweizhu, trnke,  fhieber\}@amazon.de\\
  }

\begin{document}
\maketitle
\begin{abstract}
Iterative self-refinement is a simple inference-time strategy for machine translation: an LLM revises its own translation over multiple inference-time passes. Yet document-scale refinement remains poorly understood: 1) which pipelines work best, 2) what quality dimensions improve, and 3) how refiners behave. In this paper, we present a systematic study of document-level literary translation, covering nine LLMs and seven language pairs. Across nine translation-refinement granularity combinations and five refinement strategies, we find a robust recipe: document-level MT followed by segment-level refinement yields strong and stable improvements. In contrast, document-level refinement often makes fewer edits and leads to smaller or less reliable gains. Beyond granularity, A simple general refinement prompt consistently outperforms error-specific prompting and evaluate-then-refine schemes. Our large-scale human evaluation shows that refinement gains come primarily from fluency, style, and terminology, with limited and less consistent improvements in adequacy. Experiments varying model strength reveal refinement projects outputs toward the refiner's distribution rather than performing targeted error repair. These findings clarify the mechanisms and limitations of current refinement approaches.  %

\end{abstract}

\section{Introduction}

Large language models (LLMs) have made document-level machine translation (doc-MT) increasingly practical, enabled by long-context modeling and strong generation quality~\cite{wu2024adapting,ramos2025multilingual}. Recent work further shows that multi-step, inference-time pipelines can improve translation quality~\cite{he2024exploring,briakou2024translating,tan2025remedyrgenerativereasoningmachine}. A particularly simple and widely used strategy is \emph{iterative self-refinement}~\cite{chen2024iterative,xu2024llmrefine,wu2025please}, which repeatedly revises an initial translation over multiple inference-time passes using the same LLM.

Despite growing interest, we still lack a clear understanding of refinement at the \emph{document level}. Importantly, existing refinement studies operate at a \textit{segment- or paragraph-level} granularity, even when they target document translation~\cite{briakou2024translating,wu2025please,tan2025remedyrgenerativereasoningmachine}. Additionally, these studies typically report overall translation quality, and leave three fundamental questions underexplored:
(1) \textbf{Translation pipeline design:} how should we combine translation and refinement across different granularities (segment, paragraph, document)?
(2) \textbf{Quality dimensions:} when refinement helps, does it correct meaning errors, or mainly polish fluency, style, and terminology?
(3) \textbf{Refiner behavior:} do LLM refiners resemble human post-editors that locate and minimally fix errors, or do they operate differently?

In this work, we conduct a systematic study of self-refinement for document-level translation on WMT24~\cite{kocmi2024findings}, covering nine LLMs, seven language pairs, and five refinement strategies. Additionally, we compare nine translation-refinement granularity combinations.
Across settings, we find a robust recipe: \textbf{document-level MT followed by segment-level refinement} yields strong and the most stable improvements, while document-level refinement tends to make minor edits and delivers limited gains. Surprisingly, a simple \emph{general} refinement prompt proves robust across settings, matching or exceeding more elaborate strategies such as error-specific and evaluate-then-refine schemes, raising a key question: if refinement primarily targets meaning errors, why do not adequacy-focused strategies excel? This may suggests refinement optimizes other qualities instead.

We analyze what refinement actually changes using fine-grained automatic MQM-style~\cite{freitag2021experts} evaluation that separates accuracy, fluency, style, and terminology.
We find a consistent pattern: refinement substantially improves \textbf{fluency} (and moderately improves style/terminology) but yields only limited and unreliable gains in \textbf{accuracy}.
These trends are further supported by our human evaluations: a large-scale human MQM and direct assessment study confirms the same dimension-wise pattern, while a targeted pairwise preference study shows that 98\% prefer refinement for fluency, versus much more mixed preferences for adequacy.
Further analyses of model probabilities provide evidence that refiners tend to favor more natural target-side realizations rather than more faithful source-conditioned translations.

Finally, we analyze refiner behavior by varying model strength. We find refinement exhibits a \emph{ceiling effect} (refiner strength determines the final quality upper bound) and an \emph{anchor effect} (the initial translation limits the outcome). Refiners do not target low-confidence regions, and likelihood analyses show they optimize target-side naturalness over source faithfulness. This suggests refinement operates as a projection toward the refiner's preferred distribution rather than targeted error repair—explaining the fluency gains but limited adequacy improvements. Our contributions are:

\paragraph{Document-level refinement pipelines.}
We analyze nine MT--refinement granularity combinations and five refinement strategies, and identify document-level MT followed by segment-level general refinement as a strong and robust recipe.

\paragraph{What refinement changes.}
Our large-scale human evaluations show that current MT refiners primarily improve fluency, with smaller gains in style/terminology and only limited improvements in semantic-level translation adequacy.

\paragraph{How MT refiners behave.}
By varying translator and refiner strength and analyzing edit patterns, we provide evidence that LLM refiners tend to project translations toward their own regime rather than performing targeted human-like post-editing.

\section{Related Work}
\label{sec:related_work}

\paragraph{Document-level MT and evaluation.}
Recent advances in long-context LLMs have shifted MT toward document-level tasks, such as translating literary works and reports~\cite{kocmi2024findings,semenov-etal-2025-findings,obrien-etal-2025-dochplt, appicharla2025survey}. Proprietary LLMs show strong performance on doc-MT~\cite{wang-etal-2023-document-level, kocmi-etal-2025-findings}. \citet{wu2024adapting, alabi-etal-2025-afridoc} adapt smaller LLMs to doc-MT via post-training. Nevertheless, LLMs still face challenges with length bias and content omission when translating overly long documents~\cite{hu-etal-2025-source,peng-etal-2025-investigating}. In addition, optimal pipelines for refining doc-MT outputs for improved quality are under-explored.

A further obstacle is evaluation: most trained metrics are optimized for sentence-level inputs and do not explicitly model long-range context~\cite{rei-etal-2022-comet, guerreiro-etal-2024-xcomet, tan-monz-2025-remedy}. Recently, LLM-as-a-judge has become popular, particularly in reference-free evaluations~\cite{kocmi2023gemba, maheswaran-etal-2025-taser, junczys-dowmunt-2025-gemba, sun-etal-2025-fine}; yet \citet{domhan2025same} show degraded reliability of LLMs on long documents and propose MQM-FSP to make judgment length-invariant. We therefore use MQM-FSP as a fine-grained judge and validate key trends with targeted human evaluation.

\paragraph{Inference-time scaling for MT.}
Beyond improving base translation models, a growing line of work explores inference-time methods to trade additional computation for higher translation quality.
This includes decoding-time strategies such as sampling and reranking~\cite{eikema2022sampling,zhao2024marco}, Best-of-$N$ selection~\cite{lee2021discriminative,tan2025investigating}, and structured multi-step generation that encourages intermediate reasoning or planning before producing a final translation~\cite{he2024exploring,briakou2024translating}.
These approaches aim to improve outputs without retraining, either by generating and selecting among candidates or by decomposing generation into multiple stages.

\paragraph{Refinement versus human post-editing.}
Human post-editing is often characterized as targeted error repair with minimal necessary edits. However, whether LLM refiners follow a similar mechanism remains uncertain. Existing refinement work rarely probes how outcomes depend on the relative strength of the translator and the refiner, or how edits relate to model confidence signals.

Our work complements prior refinement studies by jointly studying document-scale pipeline design, MQM analyses, and behavior probes under controlled translator--refiner settings. This allows us to characterize when refinement is most effective, which quality dimensions it reliably improves, and whether its behavior aligns with a human-like locate-and-fix mechanism.

\section{Task Definition and Experimental Setup}
\subsection{Task and Data}

We study iterative refinement for document-level literary machine translation.
Our experiments use the WMT24-Literary data~\cite{kocmi2024findings}, which consists of long-form narrative documents with rich discourse phenomena and style-sensitive expressions.
Compared to sentence-level news translation, this setting poses additional challenges: (i) documents often span thousands of tokens, requiring coherent discourse and consistent terminology; and (ii) evaluation must account for not only adequacy but also fluency and stylistic naturalness.

We conduct our experiments on \textit{seven} language pairs in WMT24-Literary (en-\{cs,de,es,ja,ru,zh\}, ja-zh). Unless otherwise specified, we operate on the official document boundaries provided by the shared task.

\subsection{Pipelines and Granularity}
\label{sec:setup_pipelines}
A refinement pipeline consists of two stages: \textbf{translation} and \textbf{refinement}.
We vary the granularity of both stages (i.e., the unit of translation/editing under full-document context): \textit{segment-level}, \textit{paragraph-level}, and \textit{document-level}.
Let $g_{\text{MT}} \in \{\text{seg}, \text{para}, \text{doc}\}$ denote the translation granularity and
$g_{\text{refine}} \in \{\text{seg}, \text{para}, \text{doc}\}$ the refinement granularity.
This yields nine combinations $(g_{\text{MT}}, g_{\text{refine}})$.

For each document, a translator model first produces an initial translation at granularity $g_{\text{MT}}$.
A refiner model then revises it at granularity $g_{\text{refine}}$.
Across all refinement settings, we condition the refiner on the \textit{full source document} and the \textit{full initial document translation}.
When refining at a finer granularity (e.g., doc$\rightarrow$seg/para), we additionally provide the \textit{local unit translation} to be edited and ask the model to output only the revised unit; revised units are then merged back into a document.

\subsection{Refinement Strategies}
\label{sec:strategies}
We evaluate five refinement prompting strategies commonly used in prior work~\cite{wu2025please}:
\textbf{(1) General refinement}, which requests improving the translation for overall quality;
\textbf{(2) Error-specific refinement}, which instructs the model to focus on particular error types (e.g., accuracy);
\textbf{(3) Evaluate-then-refine}, which first elicits a brief MQM critique and then asks for a revised translation;
\textbf{(4) Monolingual refinement}, which prompts LLMs to improve translation text quality; this eliminates potential strong source bias for
refinement;
and \textbf{(5) Step-By-Step Translation}~\cite{briakou2024translating}, which decomposes the MT task into four steps: \textit{Research}, \textit{Drafting}, \textit{Refinement}, \textit{Proofread}. All strategies share the same input information (source, initial translation, and context) and differ only in the instruction format.
See prompts in~\ref{appendix:prompts}.

\subsection{Models and Inference Configuration}
\label{sec:models}

\paragraph{Models.}
We conduct experiments on diverse LLMs of varying model families and sizes.
Our main experiments include the Qwen2.5 instruct series (14B, 32B, 72B)~\cite{qwen2025qwen25technicalreport}, Qwen3-32B/235B~\cite{yang2025qwen3technicalreport}, DeepSeek-V3~\cite{deepseekai2025deepseekv3technicalreport}, GPT-OSS-120B~\cite{openai2025gptoss120bgptoss20bmodel}, GPT-4o, and GPT-5.2.\footnote{See~\Cref{appendix:model_details} for the specific model versions, checkpoints, and deployment specifications.} This selection covers recent strong long-context models and mid-sized models, enabling controlled analyses of translator--refiner strength interactions.

\paragraph{Granularity and context length.}
We consider three levels of granularity for both translation and refinement:
\textbf{segment} (typically a single sentence, approximately 20--50 words),
\textbf{paragraph} (several sentences, approximately 200 words),
and \textbf{document} (approximately 2,048 words). Documents exceeding 2,048 words are split into contiguous chunks of approximately 2,048 words. See details on data processing in Appendix~\ref{appendix:data_processing_details}%

\paragraph{Decoding and refinement iterations.}
Unless otherwise specified, we use a fixed decoding configuration for all models (temperature=0).
For iterative refinement, we run up to four refinement steps, where each step takes the previous translation and produces a revision under the same strategy prompt.

\begin{table*}[t]
\centering
\scriptsize
\setlength{\tabcolsep}{2.2pt}
\renewcommand{\arraystretch}{0.92}
\resizebox{\textwidth}{!}{%
\begin{tabular}{l c cccc cccc cccc}
\toprule
& \multicolumn{1}{c}{Doc-MT}
& \multicolumn{4}{c}{Doc$\rightarrow$Seg ($\Delta$)}
& \multicolumn{4}{c}{Doc$\rightarrow$Para ($\Delta$)}
& \multicolumn{4}{c}{Doc$\rightarrow$Doc ($\Delta$)} \\
\cmidrule(lr){2-2}\cmidrule(lr){3-6}\cmidrule(lr){7-10}\cmidrule(lr){11-14}
Model & Init
& s1 & s2 & s3 & s4
& s1 & s2 & s3 & s4
& s1 & s2 & s3 & s4 \\
\midrule

GPT-5.2 & 90.0 & \gain{1.3} & \gain{2.3} & \gain{1.8} & \gain{2.0} & \gain{2.1} & \gain{2.6} & \gain{1.9} & \gain{2.2} & \gain{2.6} & \gain{2.6} & \gain{2.5} & \gain{2.3} \\
\textit{Edit ratio (\%)} & - & 27.8 & 28.4 & 31.2 & 30.5 & 27.7  & 29.3  & 30.6  & 30.7 & 25.8  & 25.5  & 25.4  & 25.5 \\
\midrule

GPT-4o & 86.4 & \gain{2.5} & \gain{2.5} & \gain{2.1} & \gain{1.9} & \gain{-0.9} & \gain{-0.5} & \gain{0.7} & \gain{1.9} & \gain{-0.9} & \gain{0.7} & \gain{-0.8} & \gain{-0.2} \\
\textit{Edit ratio (\%)} & - & 17.7 & 20.0 & 22.2 & 22.4 & 12.5  & 13.8  & 15.0  & 15.3 & 1.2  & 1.2  & 1.2  & 1.2 \\
\midrule

DeepSeek-V3-671B & 85.7
& \gain{1.6} & \gain{4.0} & \gain{4.1} & \gain{3.6}
& \gain{2.2} & \gain{1.5} & \gain{1.8} & \gain{2.4}
& \gain{0.7} & \gain{1.7} & \gain{1.7} & \gain{1.6} \\
\textit{Edit ratio (\%)} & - & 25.4 & 27.6 & 29.9 & 29.6 & 8.0  & 8.4  & 8.6  & 8.6 & 2.3  & 3.1  & 1.6  & 1.6 \\
\midrule

Qwen3-235B & 83.0
& \gain{1.0} & \gain{1.5} & \gain{2.2} & \gain{2.6}
& \gain{2.5} & \gain{4.6} & \gain{5.1} & \gain{5.7}
& \gain{3.4} & \gain{4.3} & \gain{5.1} & \gain{4.7} \\
\textit{Edit ratio (\%)} & - & 27.9 & 30.8 & 32.7 & 33.9 & 25.2 & 26.7 & 27.2 & 27.5 & 18.2 & 18.5 & 18.6 & 18.6 \\
\midrule

GPT-OSS-120B & 74.2
& \gain{4.4} & \gain{5.6} & \gain{5.3} & \gain{5.8}
& \gain{4.3} & \gain{4.5} & \gain{5.7} & \gain{6.0}
& \gain{6.5} & \gain{6.7} & \gain{7.3} & \gain{8.1} \\
\textit{Edit ratio (\%)} & -  & 34.7 & 38.9 & 41.7 & 42.6 & 32.6 & 42.6 & 47.0 & 49.9 & 17.4 & 19.0 & 22.4 & 23.0 \\
\midrule

Qwen3-32B & 70.5
& \gain{2.7} & \gain{3.3} & \gain{3.5} & \gain{3.5}
& \gain{4.4} & \gain{4.9} & \gain{5.1} & \gain{6.2}
& \gain{1.6} & \gain{1.8} & \gain{1.7} & \gain{1.5} \\
\textit{Edit ratio (\%)} & - & 31.8 & 36.3 & 38.7 & 39.0 & 13.6 & 14.4 & 14.9 & 15.0 & 1.1 & 1.1 & 1.1 & 1.1 \\
\midrule

Qwen2.5-72B & 72.7
& \gain{4.6} & \gain{6.6} & \gain{7.6} & \gain{7.0}
& \gain{3.2} & \gain{5.2} & \gain{6.0} & \gain{5.3}
& \gain{2.9} & \gain{4.0} & \gain{3.5} & \gain{2.8} \\
\textit{Edit ratio (\%)} & - & 21.6 & 24.3 & 26.2 & 26.5 & 13.2 & 17.3 & 17.3 & 17.5 & 5.8  & 5.7  & 5.7  & 5.7 \\
\midrule

Qwen2.5-32B & 64.4
& \gain{6.4} & \gain{7.0} & \gain{6.9} & \gain{6.1}
& \gain{5.7} & \gain{6.0} & \gain{6.3} & \gain{6.2}
& \gain{1.0} & \gain{-0.2} & \gain{0.8} & \gain{1.1} \\
\textit{Edit ratio (\%)} & - & 24.9 & 30.0 & 31.6 & 33.8 & 10.3 & 13.0 & 14.7 & 17.0 & 3.0  & 3.0  & 3.0  & 3.0 \\
\midrule

Qwen2.5-14B & 64.8
& \gain{0.3} & \gain{2.1} & \gain{1.3} & \gain{2.8}
& \gain{1.7} & \gain{1.3} & \gain{1.6} & \gain{0.3}
& \gain{0.8} & \gain{1.5} & \gain{1.9} & \gain{0.5} \\
\textit{Edit ratio (\%)} & - & 25.1 & 32.5 & 35.4 & 38.7 & 13.1 & 19.2 & 22.4 & 27.3 & 2.4  & 2.5  & 2.5  & 2.5 \\

\bottomrule
\end{tabular}}

\caption{Refinement performance under the \textbf{general} prompt on top of document-level MT.
Cells show MQM-FSP gains ($\Delta$) across four iterative refinement rounds (s1–s4) relative to the initial Doc-MT ($\Delta =$ refined $-$ initial), and are colored by gain tiers (\tierneg{} / \tiersmall{} / \tiermed{} / \tierlarge{}).
\textit{Edit ratio} (\%) is the fraction of target tokens that differ from the initial Doc-MT output.}
\label{tab:general_refinement_docMT}
\end{table*}

\subsection{Automatic Evaluation: Fine-grained MQM}
\label{sec:mqm_eval}

Evaluating document-level literary MT is challenging: reference-based metrics can be brittle under paraphrasing and style variation, while document-scale human MQM is costly.
We primarily rely on MQM-FSP~\cite{domhan2025same} (\texttt{Claude-3.5 Sonnet v2}), a reference-free MQM-style judge that assigns GEMBA severities~\cite{kocmi2023gemba} to MQM errors and yields dimension-wise scores (accuracy/fluency/style/terminology), aggregated to the document and system levels. Note that our MQM score is normalized to 0-100 range, covering all severity-weighted errors, and it is not the average of the dimension scores (see~\ref{appendix:automatic_eval}).

As complementary diagnostics for omissions and gross content loss, we also report d-BLEU. We keep the judge model and prompts fixed across all conditions; see Appendix~\ref{appendix:automatic_eval} for details.

\section{Refinement Pipelines at Document Scale}
\label{sec:pipelines}

We study document-scale refinement pipelines by varying (i) the granularity of the initial translation and (ii) the granularity and prompting strategy of the refinement stage.
We consider three granularities for both stages: segment, paragraph, and document.
Combining them yields nine configurations, denoted as \texttt{seg$\rightarrow$seg}, \dots, \texttt{doc$\rightarrow$doc}.

\subsection{Does Document-level MT Help?}
\label{sec:doc_mt_help}

\begin{table}[h!]
\centering
\scriptsize
\setlength{\tabcolsep}{2pt}
\renewcommand{\arraystretch}{1.35}
\resizebox{\columnwidth}{!}{%
\begin{tabular}{lccc|ccc}
\toprule
& \multicolumn{3}{c|}{MQM-FSP} & \multicolumn{3}{c}{d-BLEU} \\
Model & Seg. & Para. & Doc. & Seg. & Para. & Doc. \\
\midrule
GPT-5.2        & 83.4 & 84.7 & \textbf{90.0} & 34.5 & 33.9 & \textbf{35.9} \\
GPT-4o        & 81.1 & 84.5 & \textbf{86.4} & 38.2 & \textbf{38.2} & 37.2 \\

DeepSeek-V3        & 80.5 & 83.8 & \textbf{85.7} & 35.5 & 36.2 & \textbf{36.8} \\
Qwen3-235B & 76.4 & 82.5 & \textbf{83.0} & 32.0 & 30.6 & \textbf{34.4} \\
GPT-OSS-120B       & 62.9 & 67.2 & \textbf{74.2} & 32.5 & 28.9 & \textbf{32.1} \\

Qwen2.5-72B-it     & 68.3 & \textbf{72.8} & 72.7 & 33.4 & \textbf{33.5} & 32.9 \\

Qwen3-32B       & 58.7 & 65.5 & \textbf{70.5} & 30.6 & 31.3 & \textbf{32.4} \\

Qwen2.5-32B-it     & 56.1 & \textbf{64.8} & 64.4 & 31.0 & \textbf{31.4} & 30.7 \\

Qwen2.5-14B-it     & 47.9 & 60.7 & \textbf{64.8} & 29.5 & \textbf{31.3} & 30.4 \\
\bottomrule
\end{tabular}}
\caption{Translation granularity comparison on WMT24-Literary.
We report MQM-FSP and d-BLEU.}%
\label{tab:mt_granularity}
\end{table}

We first compare translation quality when translating at different granularities.
Table~\ref{tab:mt_granularity} reports MQM-FSP and d-BLEU for segment-, paragraph-, and document-level MT.
Across models, document-level MT is consistently strong and often best, yielding clear improvements over segment-level MT and remaining competitive with paragraph-level MT.
This suggests that, even in a purely inference-time setting, providing broader document context during generation helps models resolve local ambiguities and produce more coherent, stylistically consistent literary translations.

\smallskip\noindent\textbf{Takeaway:} \emph{In document-level literary MT, translating with full document context is a strong default that improves overall quality across model families.}

\subsection{Refinement Granularity Analysis}
\label{sec:ref_granularity}

Next, we evaluate all nine MT$\rightarrow$refinement granularity combinations under the same \textbf{general} prompt.
We center our analysis on refinement with Doc-MT (Table~\ref{tab:general_refinement_docMT}), and provide the full results in Table~\ref{tab:appendix:general_all}.

\paragraph{How much does refinement rewrite?}

As shown in Table~\ref{tab:general_refinement_docMT}, we find that the edit ratio differs strongly across refinement granularities.
Segment-level refinement rewrites the most, typically changing \textbf{25--40\%} of initial-MT tokens. Interestingly, document-level refinement is usually far more conservative for most models, with edit ratios frequently below \textbf{5\%}. One notable exception is GPT-OSS-120B, where document-level refinement is more aggressive (17--23\% edits).

\paragraph{How does refinement granularity affect performance?}

Table~\ref{tab:general_refinement_docMT} compares MQM-FSP performance under the same \textbf{general} refinement based on Doc-MT. We show that segment-level refinement yields robust improvements for most models across steps. In contrast, document-level refinement tends to have limited or unstable improvements, which seems to correlate with the low edit ratios. 

\medskip\noindent\textbf{Takeaway:}
\emph{Refinement granularity affects edit ratio and performance drastically, with segment-level refinement being the most robust choice.}

\subsection{Refinement Strategy Comparison}
\label{sec:ref_strategies}

Beyond granularity, we compare several refinement strategies applied to document-level MT outputs.
Table~\ref{tab:strategies_ds} reports DeepSeek-V3 as a representative example; See full results in Appendix~\ref{appendix:full_results}.

{%
\renewcommand{\gainbox}[2]{%
  \cellcolor{#1}\hspace{2pt}\strut #2\hspace{2pt}%
}

\begin{table}[H]
\centering
\scriptsize
\setlength{\tabcolsep}{1.0pt}
\renewcommand{\arraystretch}{1.25}

\resizebox{\columnwidth}{!}{%
\begin{tabular}{l c c c c}
\toprule
Setting & Overall & Accuracy & Fluency & Style+Term  \\
\midrule
General            & \textbf{\gain{3.6}}  & \gain{-0.2} & \textbf{\gain{2.6}}  & \textbf{\gain{1.2}} \\
Monolingual        & \gain{-0.8} & \gain{-3.8} & \gain{2.0}  & \gain{0.9} \\
Step-by-step       & \gain{1.4}  & \gain{-0.9} & \gain{-1.3} & \gain{-0.4} \\
Eval-refine        & \gain{3.5}  & \gain{-0.4} & \gain{1.5}  & \gain{0.4} \\
ErrorSpec-Accuracy & \gain{2.3}  & \textbf{\gain{1.0}}  & \gain{1.3}  & \gain{0.0} \\
ErrorSpec-Fluency  & \gain{2.0}  & \gain{0.2}  & \gain{1.6}  & \gain{0.3} \\
\bottomrule
\end{tabular}}
\caption{Relative MQM-FSP gains of refinement strategies for DeepSeek-V3 under document-level initial MT. Cells show score changes from the initial translation. Overall is not the average of dimension scores (see~\ref{appendix:automatic_eval}).}
\label{tab:strategies_ds}
\end{table}
}%

\begin{figure*}[h]
\centering
\includegraphics[width=\textwidth]{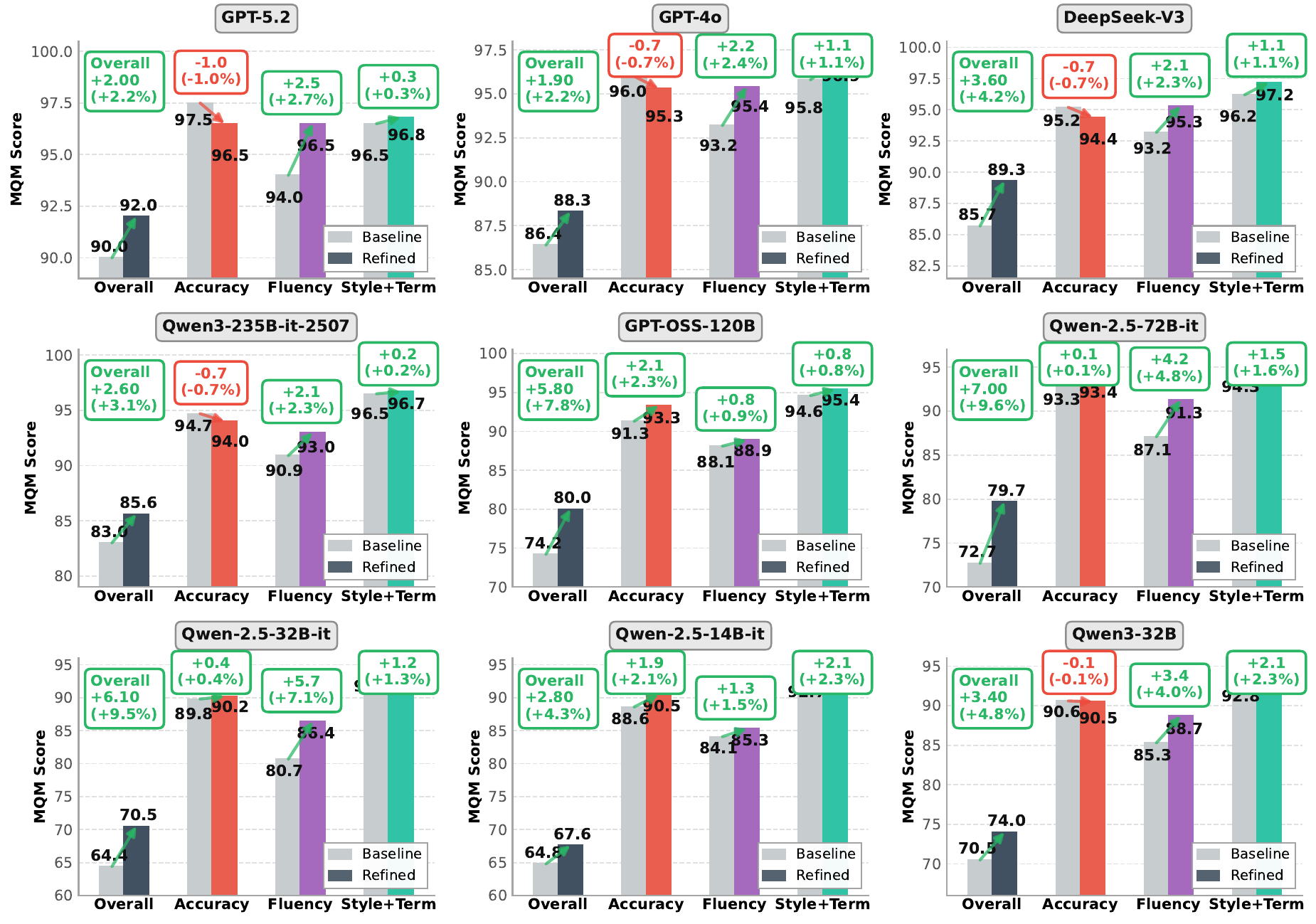}
\caption{Dimension-wise MQM-FSP gains under the Doc-Seg refinement configuration (step=4) on WMT24-Literary. $\Delta$ = refined $-$ initial; higher is better. We decompose quality into accuracy, fluency, and style+terminology. Across models and language pairs, gains are primarily driven by \textbf{fluency}, with limited \textbf{accuracy} improvement.}
\label{fig:mqm_dimensions}
\end{figure*}

\paragraph{General refinement.}
As the simplest strategy, general refinement merely asks the model to produce a better version of the current translation.
Despite this minimal instruction, it delivers the most robust improvements in our study.
We find that this simple strategy encourages edits that are less likely to introduce meaning drift, while polishing the target text.
This is also consistent with our later analysis: refinement gains are dominated by fluency and naturalness, with comparatively limited accuracy improvement (Figure~\ref{fig:mqm_dimensions}).
As a result, general refinement serves as a strong default and a high bar that more structured prompting needs to justify.

\paragraph{Monolingual rewriting.}
We include monolingual rewriting as a test of a simple motivation: if refinement mainly improves target-side readability, then removing the source might still yield fluency gains.
Although our prompt explicitly asks the model to rewrite the translation \emph{without changing its meaning}, we still observe substantial drift, reflected in frequent Accuracy drops (e.g., Table~\ref{tab:strategies_ds}).
Its Fluency improvements are often comparable to general refinement, which suggests that much of refinement’s reliable gain comes from target-side polishing.
This makes general refinement preferable in practice: it achieves similar fluency gains while keeping edits better anchored to the source.

\paragraph{Step-by-step prompting.}
Motivated by the multi-stage step-by-step MT framework of~\citet{briakou2024translating}, we evaluate a structured prompting recipe that decomposes refinement into intermediate stages instead of a single rewrite.
Compared to general refinement, it adds explicit steps like \emph{research} before editing and a final \emph{proofread}.
In our setting, step-by-step prompting underperforms the simple general strategy across all models, and our human evaluation verifies this strategy can even have a negative impact (see Table~\ref{tab:human-mqm-da-delta}).

\paragraph{Error-specific refinement.}
This approach is designed to test whether explicitly targeting a subset of MQM issues can steer refinement toward the intended dimension (e.g., emphasizing adequacy vs.\ fluency).
In our experiments, these prompts do change the profile of improvements: for instance, ErrorSpec-Accuracy yields the largest Accuracy gain on DeepSeek-V3 (Table~\ref{tab:strategies_ds}), and the fluency-focused variant tends to help Fluency.
However, these targeted gains do not consistently translate into larger Overall improvements than the general prompt across model families.
This suggests that focusing edits on one error type can leave other issues untouched and may introduce dimension trade-offs, so we treat ErrorSpec as a controllable option rather than a default strategy.

\paragraph{MQM-guided eval--refine.}
MQM-guided eval--refine makes refinement \emph{diagnosis-driven}: the model first lists MQM-style errors in the current translation and then revises accordingly.
It is often competitive, but the effect can fluctuate across refinement steps rather than improving steadily.
For example, on DeepSeek-V3 the Overall gain peaks at step2 (+4.0) and then slightly drops at step3 (+3.4) and step4 (+3.5); on Qwen3-235B it fluctuates across steps (-0.6 / +2.8 / +1.7 / +3.5) (Table~\ref{tab:doc-seg_all_1}).
We further compare the self-diagnosis errors against MQM-FSP, and find limited fine-grained overlap, especially under strict MQM style-matching (Appendix~\ref{app:eval_refine_overlap}).
Overall, the model’s self-diagnosis remains unreliable and can sometimes lead to unnecessary edits; improving self-diagnosis reliability is a promising direction for future work. 

\paragraph{Takeaway recipe.} \textit{Doc-Translation $\rightarrow$ Seg-Refinement using a simple general prompt pipeline is the most robust choice across model families and language pairs.}

\section{What Does Refinement Improve?}
\label{sec:what_changes}

\begin{table*}[t]
\resizebox{\textwidth}{!}{%
\setlength{\tabcolsep}{3pt}
\renewcommand{\arraystretch}{1.5}
\begin{tabular}{l cc|ccc || cc|ccc || cc|ccc || cc|ccc}
\toprule
& \multicolumn{5}{c||}{\textbf{OSS-120B}}
& \multicolumn{5}{c||}{\textbf{Qwen3-235B}}
& \multicolumn{5}{c||}{\textbf{DeepSeek-V3-671B}}
& \multicolumn{5}{c}{\textbf{GPT-5.2}} \\
\cmidrule(lr){2-6}\cmidrule(lr){7-11}\cmidrule(lr){12-16}\cmidrule(lr){17-21}
\textbf{Strategy}
& MQM & DA & Acc$^{\text{M}}$ & Flu$^{\text{M}}$ & S+T$^{\text{M}}$
& MQM & DA & Acc$^{\text{M}}$ & Flu$^{\text{M}}$ & S+T$^{\text{M}}$
& MQM & DA & Acc$^{\text{M}}$ & Flu$^{\text{M}}$ & S+T$^{\text{M}}$
& MQM & DA & Acc$^{\text{M}}$ & Flu$^{\text{M}}$ & S+T$^{\text{M}}$ \\
\midrule
Initial
& 19.1 & 69.3 & 58.8 & 84.4 & 76.4
& 41.8 & 75.5 & 73.8 & 88.8 & 79.7
& 45.6 & 78.4 & 72.9 & 91.4 & 81.7
& 44.8 & 77.5 & 77.5 & 92.0 & 75.6 \\
\midrule
General
& \textbf{+12.2} & \textbf{+3.7} & +3.0 & \textbf{+4.5} & \textbf{+3.7}
& \textbf{+4.4} & \textbf{+1.3} & -2.3 & +2.2 & \textbf{+4.6}
& \textbf{+7.8} & \textbf{+3.1} & -0.6 & \textbf{+1.6} & \textbf{+6.6}
& \textbf{+5.6} & \textbf{+2.0} & -1.2 & +0.7 & \textbf{+6.5} \\
 
Eval-refine
& +8.4 & +2.4 & \textbf{+4.4} & +3.0 & +1.3
& +3.3 & +0.4 & \textbf{-1.5} & \textbf{+2.6} & +2.0
& +1.7 & +0.7 & \textbf{-0.4} & +0.3 & +1.7
& +5.5 & +1.5 & \textbf{-0.6} & \textbf{+1.5} & +5.0 \\
 
Step-by-step
& -3.3 & -5.0 & -6.0 & +2.4 & +1.4
& -1.5 & -0.1 & -2.1 & +0.5 & -0.1
& -4.2 & -1.6 & -3.8 & +0.7 & -1.0
& -7.2 & -1.7 & -3.7 & -1.5 & -1.8 \\
\bottomrule
\end{tabular}%
}
\centering
\caption{Large-scale human MQM and DA evaluation results shown as score deltas relative to the \textbf{Initial} translation. Positive values indicate improvements over the initial translation. MQM denotes human MQM Overall, DA denotes Direct Assessment, and Acc$^{\text{M}}$, Flu$^{\text{M}}$, and S+T$^{\text{M}}$ denote the human MQM dimension scores for Accuracy, Fluency, and Style+Terminology, respectively. \textbf{Bold} indicates the largest improvement for each model and metric.}
\label{tab:human-mqm-da-delta}
\end{table*}

\begin{table*}[h!]
\centering
\small
\setlength{\tabcolsep}{8pt}
\renewcommand{\arraystretch}{1.08}
\begin{tabular}{lccc|c|cc|c}
\toprule
& \multicolumn{3}{c|}{Preference (\%)} & Win (\%)* & \multicolumn{2}{c|}{Win by LP (\%)*} & Sig. \\
Dimension & Init & Tie & Refined & (no ties) & en$\rightarrow$de & en$\rightarrow$es &  \\
\midrule
Accuracy
& 15.7 & 34.3 & 50.0
& 76.1
& 51.9 (14/27) & 92.5 (37/40)
& $p<10^{-4}$ \\
Fluency
& 2.0 & 21.6 & 76.5
& 97.5
& 97.0 (32/33) & 97.9 (46/47)
& $p<10^{-4}$ \\
Style+Term
& 4.9 & 29.4 & 65.7
& 93.1
& 89.3 (25/28) & 95.5 (42/44)
& $p<10^{-4}$ \\
\bottomrule
\end{tabular}
\caption{Human pairwise preferences comparing DeepSeek-V3 initial translations vs refined translations (step 4) on 200--300 word chunks.
A/B presentation is randomized per chunk.
\textbf{Win (\%)} excludes ties (*). Counts are Refined wins / non-tie comparisons.}
\label{tab:human_pref}
\end{table*}

Section~\ref{sec:pipelines} shows that refinement is highly effective under \texttt{doc$\rightarrow$seg} with a simple general prompt, while more targeted strategies do not reliably help.
We now ask a more diagnostic question: \emph{which quality dimensions does refinement actually improve?}
To answer this, we decompose quality into MQM-style dimensions and analyze refinement effects along accuracy, fluency, style+terminology.

Figure~\ref{fig:mqm_dimensions} shows a clear pattern across models and language pairs: refinement gains are driven primarily by fluency, with smaller gains in style+terminology, while accuracy improvements are weaker and less consistent. Because automatic judging can still be imperfect at document scale, we further validate these trends with complementary human evaluations.

\subsection{Large-scale Human MQM and DA Evaluation}
\label{sec:human_large_scale}

\paragraph{Setups} We commissioned professional translation vendors to assess the full WMT24-Literary test set across 7 language directions, covering 16 systems (4 LLMs $\times$ 4 strategies) and more than 1 million translated words. Annotators provided both MQM-based assessments and a holistic Direct Assessment (DA) score on a 0--100 scale reflecting their overall impression (details in Appendix~\ref{appendix:human_eval}). 

\paragraph{Results} Table~\ref{tab:human-mqm-da-delta} summarizes the resulting score deltas relative to the initial translation, including human MQM overall, DA, and MQM dimension scores. Across all four models, \textit{General} yields the largest improvements in human MQM overall and DA, \textit{Eval-refine} is typically second, and \textit{Step-by-step} performs worst. The dimension-wise MQM results are broadly consistent with our automatic analyses: the most reliable gains come from target-side improvements, especially fluency and style/terminology, whereas accuracy improvements are weaker and less consistent across models.

\subsection{Pairwise Human Preference Evaluation}
\label{sec:human_pairwise}

\paragraph{Setups} Beyond the large-scale human MQM and DA evaluation above, we also conduct a targeted pairwise human study on DeepSeek-V3, comparing the initial document-level translation against its refined output at step 4 on 200--300 word chunks from en$\rightarrow$de and en$\rightarrow$es. Professional translators provide 5-point pairwise preferences with randomized A/B order (\textit{A is much better / A is slightly better / tie / B is slightly better / B is much better}). We evaluate three dimensions: accuracy, fluency, and style+terminology. Full guidelines and the annotation interface are provided in Appendix~\ref{appendix:human_eval}.

\paragraph{Results} As shown in Table~\ref{tab:human_pref}, refined translations are strongly preferred on fluency, style, and terminology, while accuracy preferences are more mixed and accompanied by a higher tie rate. This is again consistent with the automatic MQM analysis and the large-scale human MQM/DA results.

\medskip\noindent\textbf{Takeaway:} \emph{Across both automatic MQM-FSP and human evaluation, refinement benefits are strongest in fluency and style-related dimensions, whereas accuracy improvements are weaker and less consistent. This suggests that current refinement works primarily as target-side polishing rather than reliable meaning repair.}\smallskip

This motivates the next question: \emph{does refinement behave like human post-editing (locate-and-fix) or more like broad rewriting?} Section~\ref{sec:refiner_behavior} tests this with strength-swapping and edit-location probes.

\section{How Do Refiners Behave?}
\label{sec:refiner_behavior}

\begin{table*}[t]
\centering
\small
\setlength{\tabcolsep}{3pt}
\renewcommand{\arraystretch}{1.05}
\resizebox{1.5\columnwidth}{!}{%
\begin{tabular}{l ccccc}
\toprule
Setting & Init & Step1 & Step2 & Step3 & Step4 \\
\midrule
$T_{\text{Strong}}$ (DeepSeek-V3) & 85.7 & -- & -- & -- & -- \\
$T_{\text{Strong}}\!\rightarrow\!R_{\text{Strong}}$ & 85.7 & 87.3 (+1.6) & 89.7 (+4.0) & 89.8 (+4.1) & 89.3 (+3.6) \\
\midrule
$T_{\text{Weak}}$ (Qwen2.5-14B) & 64.8 & -- & -- & -- & -- \\
$T_{\text{Weak}}\!\rightarrow\!R_{\text{Weak}}$ & 64.8 & 65.1 (+0.3) & 66.9 (+2.0) & 66.1 (+1.3) & 67.6 (+2.8) \\
\midrule
$T_{\text{Strong}}\!\rightarrow\!R_{\text{Weak}}$ & 85.7 & 75.9 (-9.8) & 76.2 (-9.5) & 75.5 (-10.2) & 77.5 (-8.1) \\
$T_{\text{Weak}}\!\rightarrow\!R_{\text{Strong}}$ & 64.8 & 83.1 (+18.2) & 86.5 (+21.7) & 88.3 (+23.4) & 88.9 (+24.0) \\
\bottomrule
\end{tabular}}
\caption{MQM performance of Translator--Refiner (T/R) combinations. We use DeepSeek-V3 and Qwen2.5-14B as the \textit{Strong} and \textit{Weak} translator/refiner. Parentheses show MQM changes over the initial translation quality.}
\label{tab:strength_interaction}
\end{table*}

The fine-grained analysis in Section~\ref{sec:what_changes} suggests that refinement primarily optimizes target-side naturalness rather than performing targeted error fixing.
We now probe \emph{how} refiners operate by varying translator--refiner strength and analyzing edit patterns.
Our goal is to test whether refinement resembles human post-editing (targeting local errors) or instead behaves like a broader transformation toward the refiner's preferred distribution.

\subsection{Translator--Refiner Strength Interaction}
\label{sec:strength_interaction}

We study refinement outcomes under different combinations of translator and refiner strength.
Concretely, we pair translations produced by a strong translator (DeepSeek-V3) or a weak translator (Qwen2.5-14B-it)
with a strong or weak refiner, while keeping the general strategy and decoding configuration fixed.
Table~\ref{tab:strength_interaction} summarizes MQM-FSP performance over four refinement steps.

\paragraph{Ceiling effect: the refiner largely determines the attainable quality.}
A strong refiner can substantially improve a weak initial translation ($T_{\text{Weak}}\!\rightarrow\!R_{\text{Strong}}$),
raising MQM by over +18 points at the first step and continuing to improve with more refinement.
However, even aggressive refinement of weak translations does not match the quality achieved when starting from a strong translation refined by a strong model ($T_{\text{Strong}}\!\rightarrow\!R_{\text{Strong}}$),
indicating that refinement cannot fully compensate for a poor starting point.

\paragraph{Anchor effect: the initial translation remains an influential starting point.}
Applying a weak refiner to a strong translation ($T_{\text{Strong}}\!\rightarrow\!R_{\text{Weak}}$) consistently degrades quality,
pulling the output toward the weaker model's regime.
Yet these degraded outputs remain well above the weak model's own translations ($T_{\text{Weak}}$) and even its self-refinement ($T_{\text{Weak}}\!\rightarrow\!R_{\text{Weak}}$),
suggesting that refinement is biased by the refiner but still anchored to the provided initial translation.

\medskip\noindent\textbf{Takeaway:} \emph{Refinement exhibits a ceiling effect governed by refiner strength and an anchor effect governed by the initial translation, implying that refinement is not purely local error repair.}

\subsection{Refinement Is Not Confidence-Guided Editing}
\label{sec:confidence_edit}

Human post-editors are often described as locating problematic spans in a draft translation and applying targeted fixes.
To test whether LLM refiners behave similarly, we ask whether words revised during refinement are associated with lower decoding confidence in the initial translation.

\paragraph{Setup}
For each document, we obtain per-token log-probabilities and entropies from the initial translation, aggregate them to the word level (min over sub-tokens), and label each word as \textit{modified} vs.\ \textit{kept} by word-level diff between the initial and refined outputs.
We quantify predictiveness using ROC AUC (higher is better; 0.5 is random) and effect size (Cohen's $d$) between modified and kept words. We provide the results of DeepSeek-V3 here but we found the same pattern for all LLMs.

\begin{table}[H]
\centering
\small
\setlength{\tabcolsep}{4pt}
\renewcommand{\arraystretch}{1.05}
\begin{tabular}{lrrrr}
\toprule
LP & \#Words & Mod.\% & AUC$_{\text{lp}}$ & AUC$_{\text{ent}}$ \\
\midrule
en-cs & 5{,}030  & 25.1 & 0.490 & 0.491 \\
en-de & 6{,}036  & 25.0 & 0.518 & 0.516 \\
en-ja & 10{,}961 & 20.7 & 0.512 & 0.515 \\
en-zh & 6{,}668  & 13.0 & 0.517 & 0.516 \\
en-ru & 4{,}888  & 28.8 & 0.485 & 0.486 \\
en-es & 6{,}030  & 26.0 & 0.509 & 0.510 \\
ja-zh & 4{,}018  & 10.1 & 0.493 & 0.494 \\
\midrule
Avg.  & 43{,}631 & 21.2 & 0.503 & 0.504 \\
\bottomrule
\end{tabular}
\caption{Predicting whether a word will be modified by refinement using initial translation confidence proxies.
AUC$_{\text{lp}}$ uses $-\log p$ (equivalently lower log-prob indicates lower confidence), and AUC$_{\text{ent}}$ uses token entropy.}
\label{tab:conf-vs-edit}
\end{table}

\paragraph{Results} %

Across 7 language pairs, we find \textit{no evidence} that refiners preferentially edit low-confidence words: both confidence proxies are near chance at predicting edit locations (Avg.\ ROC AUC $=0.503$ for log-probability and $0.504$ for entropy).
Edits are instead distributed broadly, supporting a view of refinement as global rewriting/polishing rather than a human-like locate-and-fix mechanism.

\medskip\noindent\textbf{Takeaway:} \emph{Refiners do not systematically target low-confidence regions, diverging from targeted human post-editing.}

\subsection{A Distribution-Projection View of Refinement}
\label{sec:projection_view}

The above evidence suggests a simple picture of refinement.
Instead of locating a small set of errors and fixing them, refiners often perform broad target-side polishing: they rewrite the draft into a version that reads more naturally in the target language, while staying anchored to the provided translation (and to the source when available).
We call this a \emph{distribution-projection} view: refinement moves outputs toward the refiner’s preferred target-text distribution under an anchoring constraint.

This view is consistent with our findings: gains concentrate in fluency/style, edits are not concentrated on low-confidence regions, and the attainable quality is largely determined by the refiner (ceiling) while still shaped by the initial translation (anchor).

\paragraph{Likelihood measurement.}
We quantify this view using the refiner’s own likelihood, comparing the initial translation $y^{(0)}$ and its refined version $y^{(r)}$ under the same model $p_\theta$. Here $x$ denotes the complete source document and $y$ denotes the full document translation. We report two scores: an \emph{unconditional} target-text score $s(y)$ and a \emph{source-conditioned} score $s(y\mid x)$, and compute changes $\Delta s$ and $\Delta s_x$ from $y^{(0)}$ to $y^{(r)}$.

{%
\setlength{\abovedisplayskip}{2pt}
\setlength{\belowdisplayskip}{2pt}
\setlength{\abovedisplayshortskip}{2pt}
\setlength{\belowdisplayshortskip}{10pt}
\small
\[
s(y\mid x)=\frac{1}{|y|}\log p_\theta(y\mid x), \qquad
s(y)=\frac{1}{|y|}\log p_\theta(y).
\]
\[
\Delta s_x = s(y^{(r)}\mid x)-s(y^{(0)}\mid x), \qquad
\Delta s = s(y^{(r)})-s(y^{(0)}).
\]
}%

Table~\ref{tab:logprob_diag_dsv3} shows that refinement consistently increases the target-text score ($\Delta s>0$), while changes in the source-conditioned score are smaller and often negative ($\Delta s_x\le 0$), especially under \texttt{doc$\rightarrow$seg}.
This pattern matches the distribution-projection view: refinement reliably improves target-side naturalness, while improvements under strict source-conditioning are weaker and not guaranteed.

\begin{table}[t]
\centering
\small
\setlength{\tabcolsep}{3pt}
\renewcommand{\arraystretch}{1.05}
\resizebox{0.85\columnwidth}{!}{%
\begin{tabular}{lcccc}
\toprule
Setting & s1 & s2 & s3 & s4 \\
\midrule
\multicolumn{5}{c}{\textit{$\Delta s_x = s(y^{(r)}\mid x)-s(y^{(0)}\mid x)$}} \\
\cmidrule(lr){1-5}
doc-seg & -0.151 & -0.164 & -0.175 & -0.175 \\
doc-par & -0.051 & -0.055 & -0.057 & -0.057 \\
doc-doc & -0.001 & -0.001 & -0.001 & -0.001 \\
\midrule
\multicolumn{5}{c}{\textit{$\Delta s = s(y^{(r)})-s(y^{(0)})$}} \\
\cmidrule(lr){1-5}
doc-seg & 0.041 & 0.051 & 0.055 & 0.057 \\
doc-par & 0.035 & 0.037 & 0.038 & 0.039 \\
doc-doc & 0.004 & 0.002 & 0.007 & 0.007 \\
\bottomrule
\end{tabular}}
\caption{Internal probability diagnostics for DeepSeek-V3 self-refinement. $s(y\mid x)$ and $s(y)$ are length-normalized log-likelihoods under the same refiner model. We report deltas as refined minus initial.}
\label{tab:logprob_diag_dsv3}
\end{table}

\section{Conclusion}
\label{sec:conclusion}

We presented a systematic study of iterative refinement for document-level literary machine translation on WMT24-Literary, spanning multiple LLMs and language pairs.
Across nine translation--refinement granularity combinations and multiple prompting strategies, we identify a robust pipeline: document-level MT followed by segment-level refinement with a simple general prompt.
Our detailed human evaluations show that refinement gains are driven primarily by \textit{fluency, style, and terminology}, with only limited improvements in semantic-level translation adequacy.
Mechanistic probes further reveal ceiling and anchor effects and weak coupling between model confidence and edit locations, supporting a \textit{distribution-projection} view of refinement.
Together, our findings clarify what current refinement methods can (and cannot) reliably deliver at document scale and provide practical guidance for deploying refinement in long-form translation.

\newpage
\section*{Limitations}

\paragraph{Efficiency concerns.}
While iterative refinement is effective across models and requires no additional training, the improved performance comes at the cost of increased inference time, similar to other test-time scaling approaches~\cite{wei2022cot, deepseekai2025deepseekr1, muennighoff-etal-2025-s1}. The trade-off is justified for quality-critical applications: performing refinement with state-of-the-art LLMs achieves the best translation quality. Nevertheless, this computational overhead can be significant, particularly when translating large-scale documents. Consequently, future work can explore more efficient refinement strategies.

\section*{Potential Risks}
We acknowledge societal biases may exist in Machine Translation research. To alleviate the potential risks such as toxicity and human biases, we prioritize high-quality data from WMT Metric Shared tasks.

\section*{Acknowledgments}

Shaomu Tan would like to thank Mingyang Wang (Amazon), Miaoran Zhang (Amazon), Alayi Adalibieke for their moral support during his internship at Amazon.

Bill Byrne holds concurrent appointments as an Amazon Scholar and as Professor of Information Engineering at the University of Cambridge. This paper describes work performed at Amazon.

\bibliography{custom}

\newpage

\appendix

\section{Implementation details}\label{sec:appendix}

\subsection{Model details}\label{appendix:model_details}

We provide the model identifiers in \Cref{tab:appendix:hf_models_ids}. All Hugging Face models were deployed using SGLang~\cite{sglang} on NVIDIA H100/H200 GPUs. Claude models are called via APIs.

\begin{table}[H]
    \centering
    \small

    \begin{tabularx}{\columnwidth}{l X}
        \toprule
        \textbf{Model Family} & \textbf{Hugging Face/API ID} \\
        \midrule
        
        \multirow{2}{*}{Qwen2.5} 
            & \href{https://huggingface.co/Qwen/Qwen2.5-14B-Instruct}{\texttt{Qwen/Qwen2.5-14B-Instruct}} \\
            & \href{https://huggingface.co/Qwen/Qwen2.5-32B-Instruct}{\texttt{Qwen/Qwen2.5-32B-Instruct}} \\
        \addlinespace
        
        \multirow{2}{*}{Qwen3} 
            & \href{https://huggingface.co/Qwen/Qwen3-32B}{\texttt{Qwen/Qwen3-32B}} \\
            & \href{https://huggingface.co/Qwen/Qwen3-235B-A22B-Instruct-2507}{\texttt{Qwen/Qwen3-235B-\allowbreak A22B-\allowbreak Instruct-\allowbreak 2507}} \\
        \addlinespace
        
        DeepSeek V3
            & \href{https://huggingface.co/deepseek-ai/DeepSeek-V3}{\texttt{deepseek-ai/DeepSeek-V3}} \\
        \addlinespace
        
        GPT-OSS 
            & \href{https://huggingface.co/openai/gpt-oss-120b}{\texttt{openai/gpt-oss-120b}} \\
        \addlinespace
        
        Claude 3.5
            & \texttt{anthropic.claude-3-5-\allowbreak sonnet-\allowbreak 20241022-\allowbreak v2:0} \\
        \addlinespace

        GPT-4o
            & \href{https://developers.openai.com/api/docs/models/gpt-4o}{\texttt{gpt-4o-2024-08-06}} \\
        \addlinespace
        
        GPT-5.2
            & \href{https://developers.openai.com/api/docs/models/gpt-5.2}{\texttt{gpt-5.2-2025-12-11}} \\
            
        \bottomrule
    \end{tabularx}
    \caption{List of models with their corresponding Hugging Face and Bedrock Model IDs.}
    \label{tab:appendix:hf_models_ids}
\end{table}

\subsection{Data processing}
\label{appendix:data_processing_details}

We describe the procedure for obtaining text segments at different granularity levels and the rationale behind our segmentation strategy.

\paragraph{Data source and initial format.}
The WMT24 dataset is provided in a pre-segmented format, where the raw data is already split into segment-level units, typically corresponding to individual sentences. These segments serve as the atomic units in our processing pipeline.

\paragraph{Processing methodology.}
To construct our multi-granularity dataset, we first merge all pre-segmented units into complete documents by concatenating consecutive segments with double newline delimiters (\texttt{\textbackslash n\textbackslash n}). From these reconstructed documents, we then create three levels of granularity:

\begin{itemize}
    \item \textbf{Segment-level:} We retain the original pre-segmented units without modification. Each segment typically corresponds to a single sentence (approximately 20--50 tokens).
    
    \item \textbf{Paragraph-level:} We group consecutive segments until reaching a target length of approximately 250 words (±50 words tolerance). Each paragraph thus contains multiple complete sentences (approximately 200 tokens).
    
    \item \textbf{Document-level:} We group consecutive segments until reaching a target length of approximately 2,048 words (±500 words tolerance). Documents exceeding this range are split into multiple contiguous chunks, each adhering to the target length constraint.
\end{itemize}

Note that we do not split within a segment, ensuring that all granularity levels consist of complete, consecutive segments and preserving the linguistic integrity of individual sentences.

The rationale for defining the document limit at 2,048 words is threefold: (a) it is sufficient to capture the majority of discourse dependencies~\cite{kim:2025:WMT1}; (b) certain LLMs have generation limits that prevent generating longer outputs; and (c) \citet{kocmi-etal-2025-findings} have shown that contemporary LLMs, including strong proprietary models, suffer from significant content omission when translating very long documents.

\begin{table*}[t]
\centering
\scriptsize
\setlength{\tabcolsep}{3.2pt}
\renewcommand{\arraystretch}{1.05}
\resizebox{\textwidth}{!}{%
\begin{tabular}{l r r r c c c c c c c c c}
\toprule
Model
& \#Err$_{\text{Eval}}$
& \#Err$_{\text{FSP}}$
& Ratio
& \multicolumn{3}{c}{Error-type match (cat)}
& \multicolumn{3}{c}{MQM match (cat+sev)}
& \multicolumn{3}{c}{Span match (cat+sev+span@0.3)} \\
\cmidrule(lr){5-7}\cmidrule(lr){8-10}\cmidrule(lr){11-13}
& & & & P & R & F1 & P & R & F1 & P & R & F1 \\
\midrule
DeepSeek-V3-671B
& 1921 &  643 & 2.99
& 0.0906 & 0.2706 & 0.1357
& 0.0786 & 0.2348 & 0.1178
& 0.0396 & 0.1182 & 0.0593 \\
Qwen3-235B
& 2084 &  734 & 2.84
& 0.1036 & 0.2943 & 0.1533
& 0.0729 & 0.2071 & 0.1079
& 0.0441 & 0.1253 & 0.0653 \\
Qwen3-32B
& 3391 & 1259 & 2.69
& 0.1439 & 0.3876 & 0.2099
& 0.1153 & 0.3106 & 0.1682
& 0.0752 & 0.2025 & 0.1097 \\
Qwen2.5-14B
& 2635 & 1413 & 1.87
& 0.1977 & 0.3687 & 0.2574
& 0.1393 & 0.2597 & 0.1813
& 0.0861 & 0.1607 & 0.1122 \\
Qwen2.5-32B
& 1868 & 1454 & 1.28
& 0.1809 & 0.2325 & 0.2035
& 0.1526 & 0.1960 & 0.1716
& 0.0969 & 0.1245 & 0.1090 \\
\bottomrule
\end{tabular}%
}
\caption{Overlap between Eval--Refine \textbf{diagnosis-stage} MQM errors and MQM-FSP errors on the same initial translations.
\textbf{Error-type match}: MQM category only. \textbf{MQM match}: MQM category + severity. \textbf{Span match}: category + severity + span overlap (threshold 0.3).}
\label{tab:eval_refine_overlap_summary}
\end{table*}

\subsection{Refinement Prompts}\label{appendix:prompts}
We provide our refinement prompts as follows:
\begin{itemize}
    \item \textbf{General translation refinement}: These prompts refine translations at different granularities (segment, paragraph, and document levels) with full source context: \Cref{tab:appendix:general_segment_refinement_prompt,tab:appendix:general_paragraph_refinement_prompt,tab:appendix:general_document_refinement_prompt}.
    \item \textbf{Monolingual refinement}: This prompt refines translations without reference to the source document: \Cref{tab:appendix:monolingual_refinement_prompt}.
    \item \textbf{Step-by-step prompting:}~\cite{briakou2024translating}: A pipeline of four prompts applied sequentially, with each step addressing a specific subtask and refining the previous output to form the final translation. All four prompts are provided in \Cref{tab:appendix:stepbystep_refinement_prompt}.
    \item \textbf{Error-specific refinement}: This prompt targets specific error types (accuracy or fluency) for focused correction: \Cref{tab:appendix:error_specific_refinement_prompt}.
    \item \textbf{MQM-guided eval–refine}: This prompt uses the MQM framework to evaluate and identify translation errors systematically: \Cref{tab:appendix:mqm_refinement_prompt}.

\end{itemize}

\subsection{Automatic evaluation}\label{appendix:automatic_eval}
We use three automatic evaluation metrics to assess document-level translation quality.
\paragraph{MQM-FSP.}We use MQM-FSP~\cite{domhan2025same} as our primary evaluation metric. This approach addresses the ``length bias'' found in previous LLM-based metrics like GEMBA~\cite{kocmi2023gemba}, which often fail to detect translation errors as the input length increases. By prompting the LLM to evaluate the translation sentence by sentence while maintaining the full document context, MQM-FSP ensures the evaluator is length-invariant and improves system ranking. \Cref{tab:appendix:mqm_fsp_prompt} shows the MQM-FSP prompt.

To account for varying document lengths, we compute a length-normalized quality score for each document. First, we aggregate error spans identified in the document, applying weights ($w$) of 1, 3, and 5 for minor, major, and critical errors, respectively. We then normalize this weighted error sum to a standard basis of 1,000 tokens. The final score $s$ for a given document is calculated as:

\begin{equation*}
s = \max\left(0, 100 - \frac{\sum_{i=1}^{N} w_i}{|D|} \times 1000\right)
\end{equation*}

where $N$ denotes the total number of errors, $w_i$ represents the weight of the $i$-th error, and $|D|$ is the document length in tokens. Scores are averaged first across the documents within each language pair, and finally across all 7 language pairs to report the system-level performance (e.g., the results reported in Table 1 are derived this way).

\paragraph{Dimension scores and \textit{Overall}.}
In addition to the \textit{Overall} MQM-FSP score above, we report coarse dimension-wise scores (Accuracy, Fluency, and Style+Terminology) by summing the same severity-weighted error contributions after mapping each error’s \texttt{error\_category} to a small set of buckets.
\footnote{We use keyword matching on \texttt{error\_category} to assign errors to buckets. This heuristic mapping is \emph{not} guaranteed to cover all error categories (unmatched cases are grouped as \textit{Other}) and is \emph{not} mutually exclusive (an error may match multiple buckets and contribute to multiple dimension scores).}
By contrast, \textit{Overall} is computed as the severity-weighted sum over \emph{all} errors, followed by length normalization. Therefore, \textit{Overall} is not a simple average of the dimension scores and should not be interpreted as a decomposition of them.

\paragraph{d-BLEU.} We use document-level BLEU (d-BLEU)\footnote{d-BLEU computes the BLEU score over document pairs by treating each document as a single continuous sequence, comparing the generated translation against the reference text. } as a sanity check. While LLM-based judges are effective, they can exhibit bias toward specific writing styles and may occasionally fail to penalize content omissions. d-BLEU serves as a non-neural, reference-based metric to complement the LLM evaluation.

\paragraph{Length ratio.} We also monitor the length ratio between the translation and the source text as a safeguard against catastrophic omissions. We observed that LLMs are prone to significant content omission when processing long documents (e.g., inputs exceeding 4K tokens), a tendency also reported by \citet{kocmi-etal-2025-findings}. Although our method processes text in 2K-token segments to mitigate this risk, we retain this metric as a guardrail. Specifically, we flag translations for manual inspection if the ratio of translation length to source length falls outside the range of $[0.8, 1.2]$. In our experiments, no translations triggered this threshold.

\begin{table*}[ht]
\centering
\small
\setstretch{0.9}
\begin{promptblocksml}

You are an annotator for the quality of machine translation. Your task is to assess the overall quality of the translation and to identify specific errors using Multidimensional Quality Metrics (MQM). You will be given a full document and its translation, but only score one segment at a time which is given in <target_segment></target_segment> tags. Based on the source text (in <source></source> tags) and machine translation (in <translation></translation> tags), first explain the overall translation quality of the target segment, then assign it a score, and then identify and classify all individual errors. The categories of errors are: accuracy (addition, mistranslation, omission, untranslated text), fluency (character encoding, grammar, inconsistency, punctuation, register, spelling), style (awkward), terminology (inappropriate for context, inconsistent use), other (other). Each error, including omissions or untranslated content, is classified as one of three categories: critical, major, and minor. Critical errors inhibit comprehension of the text. Major errors disrupt the flow, but what the text is trying to say is still understandable. Minor errors are technically errors, but do not disrupt the flow or hinder comprehension. The source text must be fully covered and any omissions should also be annotated as errors. Please only include errors and no spans that do not contain errors.
Please respond in JSON following this schema:
{
  "type": "object",
  "properties": {
    "quality_explanation": {
      "type": "string",
      "description": "An explanation of the overall quality of the target segment's translation considering all of the error types. When helpful, reference specific errors."
    },
    "quality_score": {
      "type": "integer",
      "description": "Overall quality score of the target segment's translation. Considering all errors, please choose the overall quality score. The quality levels associated with numerical scores: 0: No meaning preserved: Nearly all information is lost in the translation. 33: Some meaning preserved: Some of the meaning is preserved but significant parts are missing. The narrative is hard to follow due to errors. The text may be phrased in an unnatural/awkward way. Grammar may be poor. 66: Most meaning preserved and few grammar mistakes: The translation retains most of the meaning. It may have some grammar mistakes or minor inconsistencies. 100: Perfect meaning and grammar: The meaning and grammar of the translation is completely consistent with the source. The text sounds like native text in the target language without any awkward phrases. Use any number in the range between 0 and 100 for a fine-grained quality score."
    },
    "errors": {
      "type": "array",
      "items": {
        "type": "object",
        "properties": {
          "explanation": {
            "type": "string",
            "description": "A brief explanation of the error and its impact."
          },
          "error_span": {
            "type": "string",
            "description": "The relevant input span where the error occurred."
          },
          "error_category": {
            "type": "string",
            "enum": ["accuracy", "fluency", "style", "terminology", "other"],
            "description": "The main category of the error"
          },
          "error_type": {
            "type": "string",
            "description": "The specific type of error within the category."
          },
          "severity": {
            "type": "string",
            "enum": ["critical", "major", "minor"],
            "description": "The severity level of the error."
          }
        },
        "required": ["explanation", "error_category", "error_type", "severity"]
      }
    }
  },
  "required": ["quality_explanation", "quality_score", "errors"]
}
Please score the following input:
<input>
<source_language>\{\{ src_lang \}\}</source_language>
<source>\{\{ src \}\}</source>
<target_language>\{\{ tgt_lang \}\}</target_language>
<translation>\{\{ output_seq \}\}</translation>
<target_segment>\{\{ target_segment \}\}</target_segment>
</input>
Please respond in JSON without any introduction or explanation. Only the JSON response is required. Use the full document as context while only scoring the translation segment given in <target_segment></target_segment> tags.
MQM:
\end{promptblocksml}
  \caption{The \textbf{MQM-FSP prompt} for document-level translation evaluation.}
  \label{tab:appendix:mqm_fsp_prompt}
\end{table*}

\subsection{Human Evaluation}
\label{appendix:human_eval}

We conduct two complementary human evaluations: (i) a large-scale human MQM+DA study covering the full WMT24-Literary test set, and (ii) a targeted pairwise preference study designed to directly compare initial and refined translations.

\paragraph{Large-scale human MQM and DA.}
We commissioned professional translation vendors to evaluate system outputs on the full WMT24-Literary test set across 7 language directions, covering 16 systems in total. Since all MT systems translate at the document level, but full-document annotation is cumbersome, we divide each source document into smaller contiguous chunks for human assessment. Annotators are shown one source chunk together with 4--5 candidate translations from different systems; the display order of candidates is randomized independently for each source input, and annotators are blind to the originating systems.

For each candidate translation, annotators first perform MQM-style error annotation by highlighting error spans in the target text, assigning an error type, and marking a severity level (\textit{Minor}, \textit{Major}, or \textit{Critical}). We use a lightweight taxonomy with five top-level categories: \textsc{Accuracy}, \textsc{Fluency}, \textsc{Style}, \textsc{Terminology}, and \textsc{Other}. \textsc{Accuracy} covers meaning-related errors such as mistranslation, addition/hallucination, omission, and entity/number/negation errors. \textsc{Fluency} covers violations of target-language well-formedness, including grammar, orthography, and local coherence issues. \textsc{Style} captures awkwardness, register mismatch, unnatural phrasing, and source-language interference. \textsc{Terminology} covers incorrect or inconsistent technical term usage. \textsc{Other} is reserved for residual issues such as locale conventions or markup problems.

After MQM annotation, annotators additionally provide a holistic direct assessment (DA) score on a 0--100 slider reflecting their overall impression of translation quality. They are instructed to annotate all noticeable errors, remain consistent across candidates for the same source chunk, and use the most specific category available. We aggregate these annotations into human MQM overall and dimension scores; for reporting, we merge \textsc{Style} and \textsc{Terminology} into a single \textsc{Style+Terminology} dimension.

\paragraph{Pairwise human preference evaluation.}
To more directly test which quality dimensions benefit from refinement, we conduct a second targeted human study on DeepSeek-V3, comparing the initial document-level translation against its refined output at step~4. We evaluate 200--300 word chunks from en$\rightarrow$de and en$\rightarrow$es. For each chunk, professional translators compare the two candidates in randomized A/B order and judge three dimensions separately: \textsc{Accuracy}, \textsc{Fluency}, and \textsc{Style+Terminology}. Judgments are collected on a 5-point comparative scale: \textit{A much better}, \textit{A slightly better}, \textit{tie}, \textit{B slightly better}, and \textit{B much better}.

We report win/tie/loss statistics as well as win rates after excluding ties. This pairwise study complements the large-scale MQM+DA evaluation by providing a direct dimension-wise comparison between initial and refined outputs.

\begin{figure*}
    \centering
    \includegraphics[width=\textwidth]{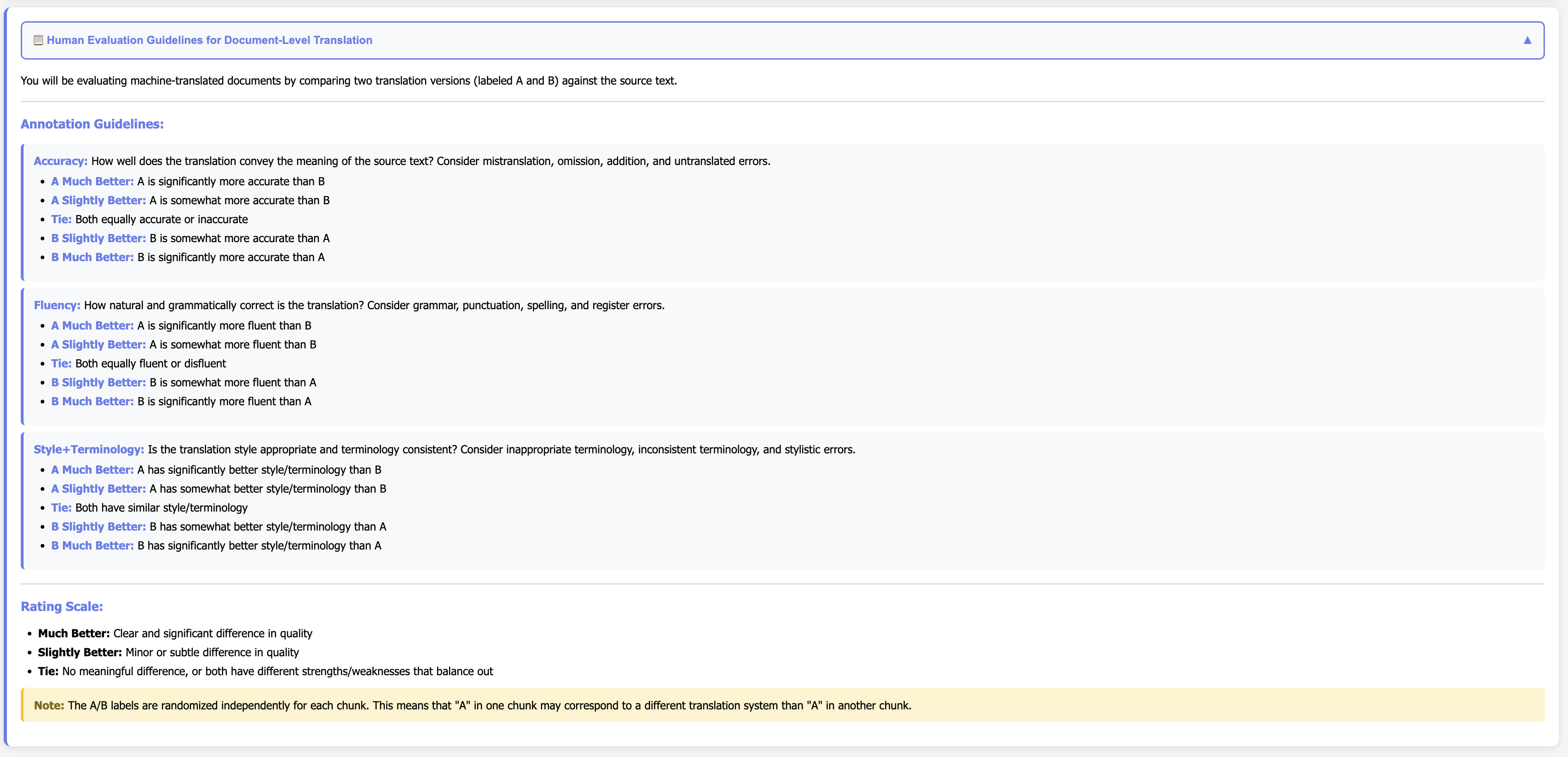}
    \caption{Annotation guidance for human evaluation.}
    \label{fig:appendix:annotation_ui_guidance}
\end{figure*}

\begin{figure*}
    \centering
    \includegraphics[width=\textwidth]{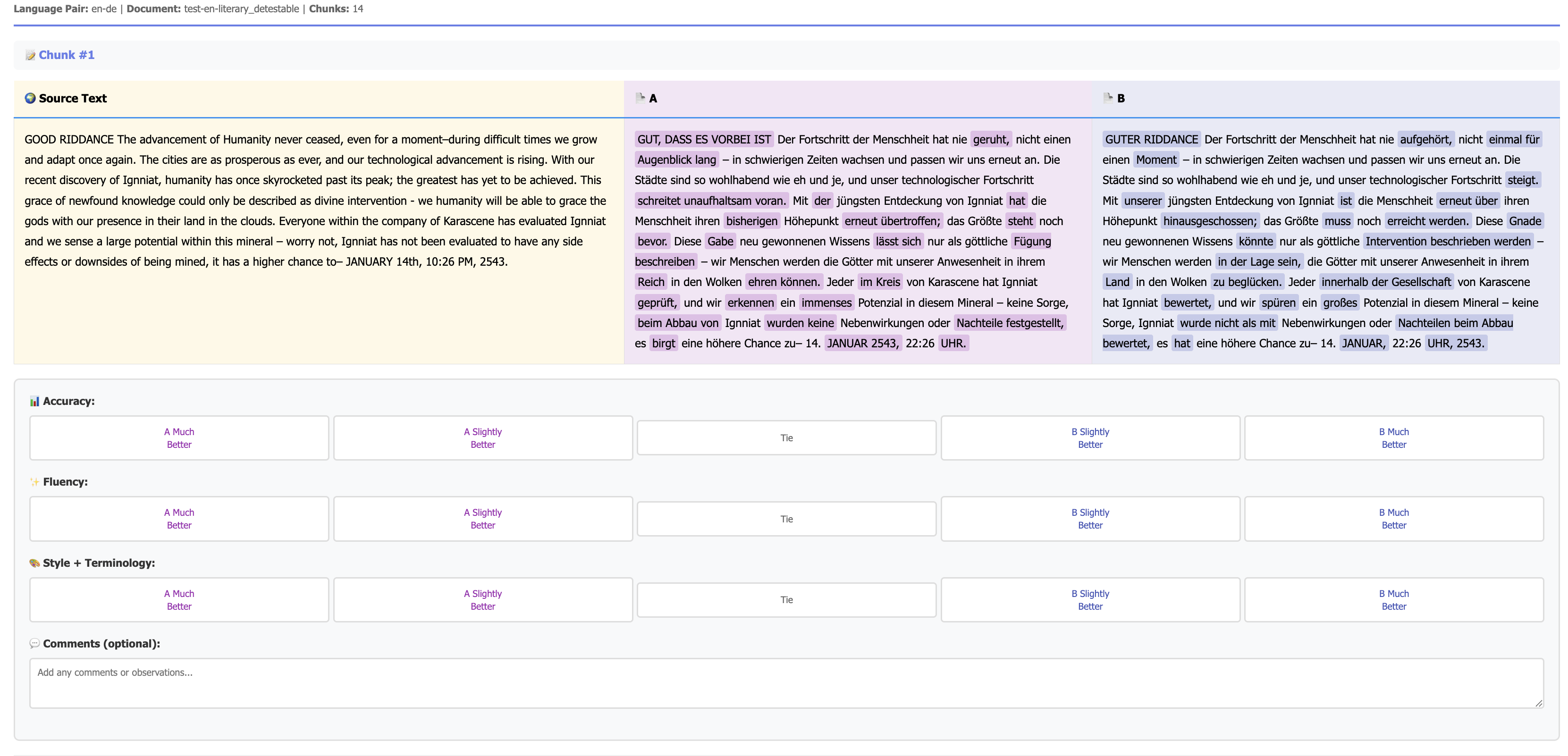}
    \caption{Example of the annotation interface for human evaluation. The source document is divided into chunks. For each chunk, annotators are shown two translations from different MT systems, with differences highlighted.}
    \label{fig:appendix:annotation_ui_example}
\end{figure*}

\begin{figure*}
    \centering
    \includegraphics[width=\textwidth]{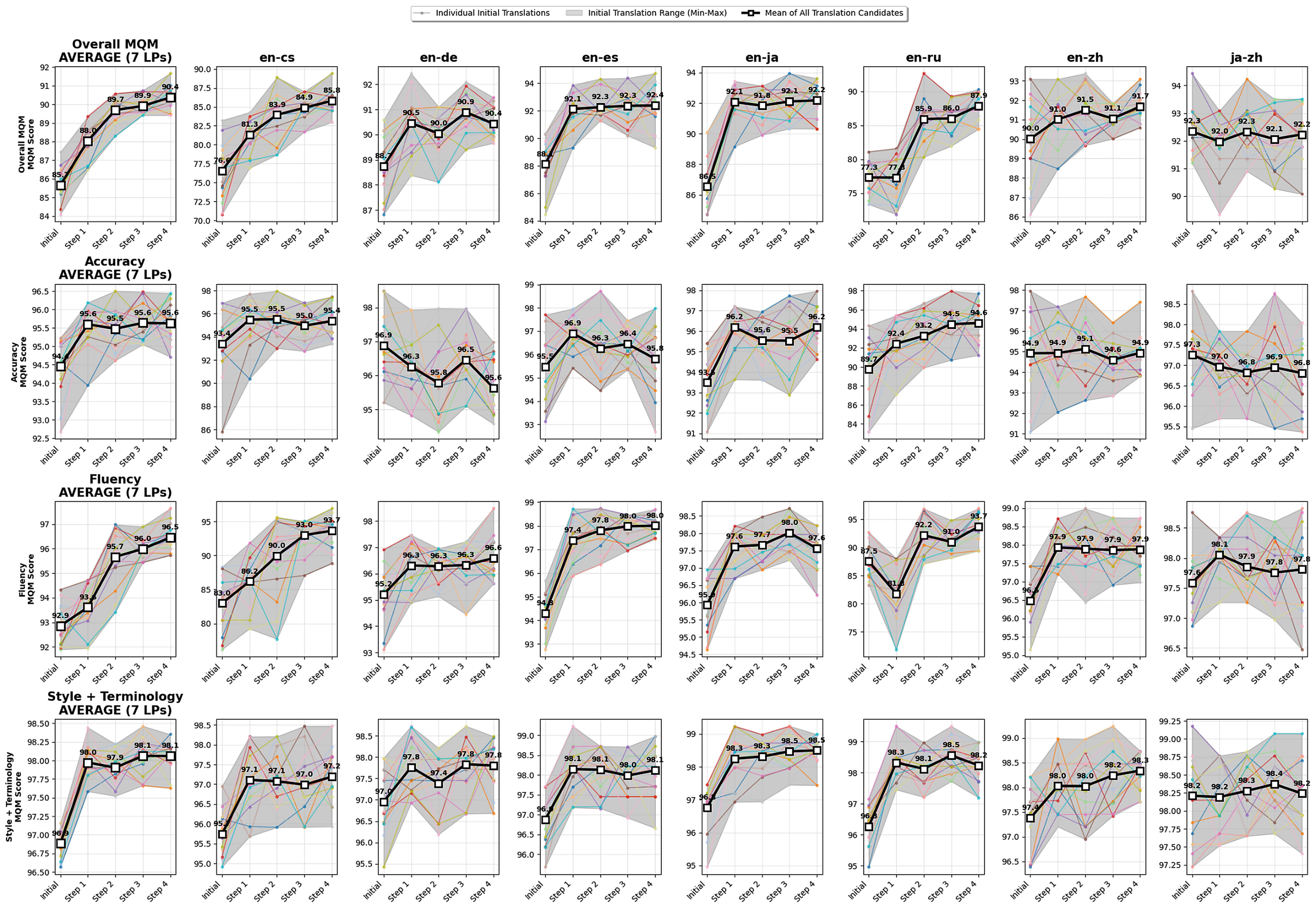}
    \caption{Refinement trajectories from 16 sampled initial document translations produced by DeepSeek-V3. Colored lines denote individual initial translations, the black line shows the candidate-wise mean, and the gray band indicates the min--max range. Across language pairs, refinement consistently improves mean MQM-FSP and reduces cross-candidate variation, especially for \textit{Overall} and \textit{Fluency}.}
    \label{fig:appendix:sampling_refine}
\end{figure*}

\section{Full Refinement Results}\label{appendix:full_results}

Due to space constraints, we provide detailed results of different refinement strategies in this section. 

\paragraph{Refinement Granularity Analysis}
We evaluate all nine combinations of \textit{initial translation} and \textit{refinement} granularities in Table~\ref{tab:appendix:general_all}. Document-level MT (docMT) consistently yields the strongest initial translations, and the same ranking persists after refinement: docMT-based refinement outperforms segment- and paragraph-based pipelines. This highlights the importance of performing document-level translation---in contrast to much prior refinement work that operates at segment/paragraph granularity even when targeting document translation.

\paragraph{Step-By-Step Translation} We demonstrate the results of Step-By-Step Translation~\cite{briakou2024translating} in Table~\ref{tab:appendix:stepbystep}. Firstly, we found this approach does not yield better performance than the \textbf{general strategy}, similar to the findings in~\citet{wu2025please}. Secondly, similar to other refinement approaches, we find that Step-by-Step does not improve translation on the \textit{Accuracy} dimension, this further reinforces our claim in Section~\ref{sec:what_changes} (i.e., refinement primarily improves fluency, while its impact on adequacy/accuracy is limited and inconsistent).

\subsection{Overlap Between Eval--Refine Diagnosis and MQM-FSP Errors}
\label{app:eval_refine_overlap}

We quantify how well the MQM error lists produced by the \textbf{diagnosis stage} of Eval--Refine align with MQM-FSP errors on the \emph{same initial translations}.
For each document, we compare two sets of error records: (i) errors predicted by Eval--Refine (diagnosis step), and (ii) errors reported by MQM-FSP (our primary judge).
We report precision/recall/F1 under three matching criteria with increasing strictness:
\textbf{Error-type match} requires MQM category agreement only;
\textbf{MQM match} requires agreement on both MQM category and severity; and
\textbf{Span match} additionally requires span-level overlap.

Table~\ref{tab:eval_refine_overlap_summary} summarizes results for all language pairs in WMT24-Literary.
Across models, Eval--Refine tends to flag more errors than MQM-FSP, and agreement decreases as we move from coarse label matching to strict span matching, suggesting the diagnosis output is better interpreted as a noisy intermediate supervision signal rather than a faithful MQM annotation.

\section{Refinement from Multiple Sampled Initial Translations}
\label{app:sampled_init_refine}

To examine whether refinement is robust to variation in the initial document translation, we sample 16 document-level initial translations from DeepSeek-V3 for each source document and apply the same iterative refinement procedure to each candidate. Figure~\ref{fig:appendix:sampling_refine} shows the MQM-FSP trajectories across refinement steps for all candidates, together with the candidate-wise mean and min--max range.

We observe two consistent patterns across language pairs. First, the mean Overall MQM-FSP score improves steadily with refinement, with the largest gains typically achieved in the first one or two steps. Second, the spread across candidates becomes noticeably smaller after refinement: while the sampled initial translations exhibit substantial variation at step~0, their scores become more concentrated after refinement. This contraction pattern is particularly clear for Overall and Fluency, while Accuracy remains comparatively stable and shows smaller, less consistent changes. Style+Terminology also tends to improve, but generally less than Fluency.

Overall, these results suggest that refinement is robust to variation in the starting translation and tends to project diverse sampled candidates toward a narrower, higher-quality region. This observation is consistent with our main finding that current refinement behaves more like target-side polishing than reliable meaning repair.

\begin{table*}[ht]
\centering
\small
\setstretch{0.9}
\begin{promptblock}
\cmtc{System prompt}
You are an expert translation quality improvement assistant. Your task is to refine a specific segment of a \{\{ tgt_lang \}\} translation to be more accurate, natural, and fluent. Output ONLY the refined segment - do NOT generate additional content.
\cmtc{User prompt}
Below is the complete \{\{ src_lang \}\} source document and its \{\{ tgt_lang \}\} translation.

**Complete Source Document (\{\{ src_lang \}\}):**
\{\{ full_src \}\}

**Complete Current Translation (\{\{ tgt_lang \}\}):**
\{\{ full_translation \}\}

-------------------------------------------------------

**Your task:** Refine segment #\{\{ segment_idx + 1 \}\} below to be:
• Accurate and faithful to the source
• Natural and fluent in \{\{ tgt_lang \}\}
• Consistent with the surrounding context
• Maintains coherence within the paragraph

**segment #\{\{ segment_idx + 1 \}\} to refine:**
\{\{ segment_to_refine \}\}
\end{promptblock}
  \caption{System and user prompt for the \textbf{\textit{segment-level} general refinement strategy}. \colorbox{commentbg}{Highlighted sections} are structural annotations for readability and are not part of the actual prompt.}
  \label{tab:appendix:general_segment_refinement_prompt}
\end{table*}

\begin{table*}[ht]
\centering
\small
\setstretch{0.9}
\begin{promptblock}
\cmtc{System prompt}
You are an expert translation quality improvement assistant. Your task is to refine a specific paragraph of a \{\{ tgt_lang \}\} translation to be more accurate, natural, and fluent.
\cmtc{User prompt}
Below is the complete \{\{ src_lang \}\} source document and its \{\{ tgt_lang \}\} translation.

**Complete Source Document (\{\{ src_lang \}\}):**
\{\{ full_src \}\}

**Complete Current Translation (\{\{ tgt_lang \}\}):**
\{\{ full_translation \}\}

-------------------------------------------------------

**Your task:** Refine paragraph #\{\{ paragraph_idx + 1 \}\} below to be:
• Accurate and faithful to the source
• Natural and fluent in \{\{ tgt_lang \}\}
• Consistent with the surrounding context
• Maintains coherence within the paragraph

**Paragraph #\{\{ paragraph_idx + 1 \}\} to refine:**
\{\{ paragraph_to_refine \}\}

Provide ONLY the improved translation without explanations.
\end{promptblock}
  \caption{System and user prompt for the \textbf{\textit{paragraph-level} general refinement strategy}. \colorbox{commentbg}{Highlighted sections} are structural annotations for readability and are not part of the actual prompt.}
  \label{tab:appendix:general_paragraph_refinement_prompt}
\end{table*}

\begin{table*}[ht]
\centering
\small
\setstretch{0.9}
\begin{promptblock}
\cmtc{System prompt}
You are an expert translation quality improvement assistant. Your task is to refine an entire \{\{ tgt_lang \}\} translation to be more accurate, natural, and fluent.
\cmtc{User prompt}
Below is a complete \{\{ src_lang \}\} source document and its \{\{ tgt_lang \}\} translation.

**Source Document (\{\{ src_lang \}\}):**
\{\{ full_src \}\}

**Current Translation (\{\{ tgt_lang \}\}):**
\{\{ full_translation \}\}

-------------------------------------------------------

**Your task:** Refine the entire translation to be:
• Accurate and faithful to the source
• Natural and fluent in \{\{ tgt_lang \}\}
• Coherent and consistent throughout
• Maintains appropriate style and tone

Provide ONLY the improved translation without explanations.
\end{promptblock}
  \caption{System and user prompt for the \textbf{\textit{document-level} general refinement strategy}.\colorbox{commentbg}{Highlighted sections} are structural annotations for readability and are not part of the actual prompt.}
  \label{tab:appendix:general_document_refinement_prompt}
\end{table*}

\begin{table*}[ht]
\centering
\small
\setstretch{0.9}
\begin{promptblock}
\cmtc{System prompt}
You are an expert translation quality evaluator. Your task is to identify and categorize translation errors using the MQM (Multidimensional Quality Metrics) framework.
\cmtc{User prompt}
Below is a complete \{\{ tgt_lang \}\} text.

**Complete Text (\{\{ tgt_lang \}\}):**
\{\{ full_translation \}\}

-------------------------------------------------------

**Your task:** Refine paragraph #\{\{ paragraph_idx + 1 \}\} below to be more natural and fluent.

**Paragraph #\{\{ paragraph_idx + 1 \}\} to refine:**
\{\{ paragraph_to_refine \}\}
Provide ONLY the improved text without explanations.\end{promptblock}
  \caption{System and user prompt for the \textbf{monolingual refinement strategy}.\colorbox{commentbg}{Highlighted sections} are structural annotations for readability and are not part of the actual prompt.}
  \label{tab:appendix:monolingual_refinement_prompt}
\end{table*}

\begin{table*}[ht]
\centering
\small
\setstretch{0.9}
\begin{promptblock}
\cmtc{Step 1 Pre-drafting Research, user prompt}
You will be asked to translate a piece of text from \{\{ src_lang \}\} into \{\{ tgt_lang \}\} following stages of the translation process. Here is the text to be translated:

Source Text: \{\{ src_text_display \}\}

To start, let's do some pre-drafting research on the above text:

Research:
During this phase, thorough research is essential to address components of the source text that pose translation challenges. The goal is to establish a comprehensive translation plan that covers the following category:

	* Idiomatic Expressions:
		* Identify idiomatic expressions that cannot be directly translated word-for-word into \{\{ tgt_lang \}\}.
\cmtc{Step 2 Drafting, user prompt}
Now, let's move on to the drafting stage.

Draft Translation:
In this phase, your primary objective is to create a draft translation that accurately conveys the meaning of the source text provided above in Step 1. At this stage, it is crucial to focus on adequacy, ensuring that your translation closely adheres to the source text.

IMPORTANT: This is a FULL TRANSLATION task, not a summary.
• Translate EVERY sentence completely
• Do NOT skip or omit any content
• Do NOT summarize or condense
• Output ONLY the translation (no notes or explanations)

Your response should conclude with the draft translation. If context is missing, generate a general translation that is adaptable to various contexts. Avoid adding any additional information not present in the source text. All elements of the source text should be present in the translation.
Provide only one best \{\{ tgt_lang \}\} translation of the \{\{ src_lang \}\} source text above, guided by the pre-drafting analysis, without adding anything further:

\{\{ tgt_lang \}\}:
\cmtc{Step 3 Refinement, user prompt}
Now let's move to the next stage.

Post-editing with local refinement:
In this stage, the primary aim is to refine the draft translation above by making micro-level improvements that improve the draft's fluency.

IMPORTANT: This is a FULL TRANSLATION task, not a summary.
• Keep EVERY sentence from the draft translation
• Do NOT skip or omit any content
• Do NOT summarize or condense

Provide only one refined \{\{ tgt_lang \}\} translation without adding anything further:

\{\{ tgt_lang \}\}:
\cmtc{Step 4 Proofread, user prompt}
Now let's move to the final stage.

Proofread and Final Editing:
The goal is to provide a polished final translation of the source text. Please refer to the source text from Step 1, the draft translation from Step 2, and the refined translation from Step 3 in the conversation above.

IMPORTANT: This is a FULL TRANSLATION task, not a summary.
• Keep EVERY sentence from the refined translation
• Do NOT skip or omit any content
• Do NOT summarize or condense

Please proofread the refined text for grammar, spelling, punctuation, terminology, and overall fluency. Ensure the translation accurately reflects the original meaning and style. Provide only the final, polished \{\{ tgt_lang \}\} translation without adding anything further:

\{\{ tgt_lang \}\}:
\end{promptblock}
  \caption{User prompts from the \textbf{step-by-step prompting} approach proposed by~\citet{briakou2024translating}. The same system prompt is used across all four translation stages: ``\texttt{You are a helpful assistant.}''. \colorbox{commentbg}{Highlighted sections} are structural annotations for readability and are not part of the actual prompt.}
  \label{tab:appendix:stepbystep_refinement_prompt}
\end{table*}

\begin{table*}[ht]
\centering
\small
\setstretch{0.9}
\begin{promptblock}
\cmtc{Refinement dimension definitions. These can be inserted as \{\{ dimension_instructions \}\}}
\cmtc{Dimension: Accuracy}
Description: Mistranslations, omissions, additions, untranslated content
Instruction:
Find ONLY accuracy errors:
  • Mistranslations (wrong meaning)
  • Omissions (missing source content)
  • Additions (extra content not in source)
  • Untranslated terms
Fix ONLY these errors. Keep all other parts UNCHANGED.
\cmtc{Dimension: Fluency}
Description: Grammar errors, awkward phrasing, unnatural word order
Instruction:
Find ONLY fluency errors:
  • Grammar mistakes (verb tense, agreement, etc.)
  • Awkward or stilted phrasing
  • Unnatural word order
Fix ONLY these errors. Keep meaning and terminology UNCHANGED.
\cmtc{System prompt}
You are an expert translation error correction specialist. Your task is to \{\{ task_description \}\} in a \{\{ target_language \}\} translation.

\{\{ dimension_instructions \}\} 
\cmtc{User prompt}
Below is the complete \{\{ source_language \}\} source document and its \{\{ target_language \}\} translation.

**Complete Source Document (\{\{ source_language \}\}):**
\{\{ full_source \}\}

**Complete Current Translation (\{\{ target_language \}\}):**
\{\{ full_translation \}\}

-------------------------------------------------------

**Your task:** Review and refine segment #\{\{ segment_number \}\} below for \{\{ focused_dimensions \}\}  errors ONLY.

IMPORTANT:
• Output ONLY the refined version of segment #\{\{ segment_number \}\}
• Do NOT include other segments
• Do NOT add explanations
• Keep the length similar to the original segment
• Identify \{\{ focused_dimensions \}\}  errors (if any) and fix ONLY those specific errors, do NOT improve other aspects

**Source segment #\{\{ segment_number \}\}:** \{\{ source_segment \}\}

**Current translation #\{\{ segment_number \}\}:** \{\{ current_segment \}\}

**Refined segment #\{\{ segment_number \}\}:**
\end{promptblock}
  \caption{Prompt template for \textbf{error-specific refinement}. The template includes dimension definitions (Accuracy and Fluency), system prompt, and user prompt with segment-level context for targeted error correction.\colorbox{commentbg}{Highlighted sections} are structural annotations 
for readability and are not part of the actual prompt.}
  \label{tab:appendix:error_specific_refinement_prompt}
\end{table*}

\begin{table*}[ht]
\centering
\small
\setstretch{0.9}
\begin{promptblock}
\cmtc{System prompt}
You are an expert \{\{ tgt_lang \}\} text quality improvement assistant. Your task is to refine a specific paragraph of a \{\{ tgt_lang \}\} text to be more natural, fluent, and coherent.
\cmtc{User prompt}
Below is the complete \{\{ src_lang \}\} source document and its \{\{ tgt_lang \}\} translation.

**Complete Source Document (\{\{ src_lang \}\}):**
\{\{ full_src \}\}

**Complete Current Translation (\{\{ tgt_lang \}\}):**
\{\{ full_translation \}\}

-------------------------------------------------------

**Your task:** Evaluate segment #\{\{ segment_idx + 1 \}\} for translation quality.

**Source segment #\{\{ segment_idx + 1 \}\}:** \{\{ src_segment \}\}

**Translation segment #\{\{ segment_idx + 1 \}\}:** \{\{ segment_to_eval \}\}

Identify all translation errors in this segment and categorize them using MQM error types:
- **Accuracy errors**: mistranslation, omission, addition, untranslated
- **Fluency errors**: grammar, spelling, punctuation, inconsistency
- **Style errors**: awkward, unnatural
- **Terminology errors**: incorrect term, inconsistent terminology

For each error, provide:
1. Error type (e.g., 'accuracy/mistranslation', 'fluency/grammar')
2. Severity (minor/major/critical)
3. The erroneous text span in the translation
4. Explanation of the error

Format your response as:
```json
{
  "errors": [
    {
      "type": "accuracy/mistranslation",
      "severity": "major",
      "text": "erroneous text",
      "explanation": "description of the error"
    }
  ],
  "overall_quality": "good/fair/poor"
}
\end{promptblock}
  \caption{System and user prompt used in the \textbf{MQM-guided eval–refine }approach.\colorbox{commentbg}{Highlighted sections} are structural annotations for readability and are not part of the actual prompt.}
  \label{tab:appendix:mqm_refinement_prompt}
\end{table*}

\begin{table*}[t]
\centering
\scriptsize
\setlength{\tabcolsep}{2.0pt}
\renewcommand{\arraystretch}{1.2}

\resizebox{\textwidth}{!}{%
\begin{tabular}{l cc cccc cccc cccc}
\toprule
& \multicolumn{2}{c}{Init MT}
& \multicolumn{4}{c}{$\rightarrow$Seg}
& \multicolumn{4}{c}{$\rightarrow$Para}
& \multicolumn{4}{c}{$\rightarrow$Doc} \\
\cmidrule(lr){2-3}\cmidrule(lr){4-7}\cmidrule(lr){8-11}\cmidrule(lr){12-15}
Model & Level & Score
& s1 & s2 & s3 & s4
& s1 & s2 & s3 & s4
& s1 & s2 & s3 & s4 \\
\midrule

\multirow{3}{*}{DeepSeek-V3}
& Seg  & 80.5
& 84.8 & 85.3 & \underline{86.3} & 85.7
& 83.9 & 85.0 & 85.5 & 85.5
& 83.0 & 83.4 & 83.3 & 83.6 \\
& Para & 83.8
& 83.4 & 84.6 & 85.3 & 86.1
& 85.7 & 86.8 & 86.9 & 87.1
& 86.9 & \underline{87.7} & 87.3 & 87.5 \\
& Doc  & 85.7
& 87.3 & 89.7 & \textbf{\underline{89.8}} & 89.3
& 87.9 & 87.2 & 87.5 & 88.1
& 86.4 & 87.4 & 87.4 & 87.3 \\
\midrule

\multirow{3}{*}{Qwen3-235B}
& Seg  & 76.4
& 82.0 & 82.7 & 82.9 & 82.4
& 84.1 & 84.8 & 84.5 & 87.6
& 87.1 & \underline{87.7} & \underline{87.7} & 87.5 \\
& Para & 82.5
& 81.9 & 83.3 & 82.8 & 82.4
& 85.0 & 85.3 & 87.2 & 86.7
& 87.7 & 88.0 & 88.2 & \underline{88.5} \\
& Doc  & 83.0
& 84.0 & 84.5 & 85.2 & 85.6
& 85.5 & 87.6 & 88.1 & \textbf{\underline{88.7}}
& 86.4 & 87.3 & 88.1 & 87.7 \\
\midrule

\multirow{3}{*}{GPT-OSS-120B}
& Seg  & 62.9
& 73.8 & 76.1 & 75.0 
& 76.0 & 74.4 & 76.5 & 78.1
& 78.4 & 77.1 & 80.6 & 77.5 & \underline{79.4} \\
& Para & 67.2
& 72.6 & 76.1 & 74.9 & 76.4
& 78.2 & 78.4 & 77.8 & 79.0
& 79.6 & 80.1 & 80.3 & \underline{81.4} \\
& Doc  & 74.2
& 78.6 & 79.8 & 79.5 & 80.0
& 78.5 & 78.7 & 79.9 & 80.2
& 80.7 & 80.9 & 81.5 & \textbf{\underline{82.3}} \\
\midrule

\multirow{3}{*}{Qwen2.5-14B}
& Seg  & 47.9
& 60.5 & \underline{65.0} & 63.6 & 61.4
& 60.9 & 63.4 & 60.9 & 57.8
& 58.9 & 57.7 & 58.1 & 58.9 \\
& Para & 60.7
& 64.9 & 66.4 & 65.4 & \underline{67.6}
& 64.2 & 63.6 & 60.6 & 57.1
& 64.0 & 64.6 & 64.2 & 64.8 \\
& Doc  & 64.8
& 65.1 & 66.9 & 66.1 & \textbf{\underline{67.6}}
& 66.5 & 66.1 & 66.4 & 65.1
& 65.6 & 66.3 & 66.7 & 65.3 \\
\midrule

\multirow{3}{*}{Qwen2.5-32B}
& Seg  & 56.1
& 66.4 & \underline{68.8} & 65.8 & 66.7
& 64.1 & 63.3 & 65.8 & 66.9
& 53.5 & 54.0 & 53.6 & 54.2 \\
& Para & 64.8
& 69.0 & 69.7 & 68.8 & 69.3
& 70.7 & 70.4 & 70.7 & \underline{70.9}
& 68.0 & 68.7 & 67.8 & 67.6 \\
& Doc  & 64.4
& 70.8 & \textbf{\underline{71.4}} & 71.3 & 70.5
& 70.1 & 70.4 & 70.7 & 70.6
& 65.4 & 64.2 & 65.2 & 65.5 \\
\midrule

\multirow{3}{*}{Qwen2.5-72B}
& Seg  & 68.3
& 73.1 & 74.8 & 76.1 & \underline{77.1}
& 73.0 & 75.2 & 76.0 & 76.7
& 74.4 & 75.2 & 75.0 & 75.4 \\
& Para & 72.8
& 76.3 & \underline{78.1} & 76.8 & 78.0
& 76.2 & 76.5 & 78.0 & 77.6
& 77.2 & \underline{78.1} & 77.7 & 76.5 \\
& Doc  & 72.7
& 77.3 & 79.3 & \textbf{\underline{80.3}} & 79.7
& 75.9 & 77.9 & 78.7 & 78.0
& 75.6 & 76.8 & 76.2 & 75.5 \\
\midrule

\multirow{3}{*}{Qwen3-32B}
& Seg  & 58.7
& 67.4 & 68.1 & 68.4 & 68.5
& 68.2 & 69.8 & 69.6 & \underline{69.9}
& 66.2 & 65.8 & 65.4 & 65.3 \\
& Para & 65.5
& 69.0 & 70.6 & 69.2 & 70.7
& 70.6 & 71.4 & 72.3 & \underline{73.0}
& 69.7 & 70.0 & 70.4 & 70.5 \\
& Doc  & 70.5
& 73.2 & 73.8 & 74.0 & 74.0
& 74.9 & 75.4 & 75.6 & \textbf{\underline{76.7}}
& 72.1 & 72.3 & 72.2 & 72.0 \\
\bottomrule
\end{tabular}%
}

\caption{MQM-FSP scores across four iterative \textbf{general refinement} rounds (s1--s4), starting from three initial levels (Seg/Para/Doc) and refining toward segment-, paragraph-, or document-level outputs. Row-wise maxima are \underline{underlined}; the best score within each model block is in \textbf{bold}.}
\label{tab:appendix:general_all}
\end{table*}

\begin{table*}[t]
\centering
\scriptsize
\setlength{\tabcolsep}{3.0pt}
\renewcommand{\arraystretch}{1.05}

\resizebox{\textwidth}{!}{%
\begin{tabular}{l l l c c c}
\toprule
Model & Setting & MQM Dim. & step2 (Draft) & step3 (Refined) & step3(Proofread) \\
\midrule

\multirow{4}{*}{GPT-5.2}
& \multirow{4}{*}{Step-by-step}
& Overall      &   89.1 & 90.6 (\gain{1.56})  & 90.5 (\gain{1.42}) \\
&  & Accuracy    & 97.1 & 97.3 (\gain{0.16}) & 97.5 (\gain{0.40}) \\
&  & Fluency     & 93.6 & 95.3 (\gain{1.74}) & 95.2 (\gain{1.68}) \\
&  & Style+Term  & 95.9 & 96.3 (\gain{0.41})  & 95.7 (\gain{-0.21}) \\
\midrule

\multirow{4}{*}{GPT-4o}
& \multirow{4}{*}{Step-by-step}
& Overall      &   83.9 & 85.6 (\gain{1.68})  & 86.0 (\gain{2.09}) \\
&  & Accuracy    & 95.9 & 96.4 (\gain{0.52}) & 96.5 (\gain{0.67}) \\
&  & Fluency     & 92.0 & 93.1 (\gain{1.07}) & 92.7 (\gain{0.69}) \\
&  & Style+Term  & 95.9 & 96.2 (\gain{0.31})  & 96.1 (\gain{0.15}) \\
\midrule

\multirow{4}{*}{DeepSeek-V3-671B}
& \multirow{4}{*}{Step-by-step}
& Overall     & 86.4 & 86.6 (\gain{0.27})  & 87.1 (\gain{0.68}) \\
&  & Accuracy    & 95.0 & 94.7 (\gain{-0.28}) & 94.8 (\gain{-0.27}) \\
&  & Fluency     & 92.7 & 92.6 (\gain{-0.07}) & 92.1 (\gain{-0.61}) \\
&  & Style+Term  & 94.9 & 95.6 (\gain{0.65})  & 96.2 (\gain{1.29}) \\
\midrule

\multirow{4}{*}{Qwen3-235B}
& \multirow{4}{*}{Step-by-step}
& Overall     & 83.9 & 84.2 (\gain{0.31})  & 84.1 (\gain{0.21}) \\
&  & Accuracy    & 93.9 & 93.3 (\gain{-0.59}) & 93.3 (\gain{-0.55}) \\
&  & Fluency     & 90.7 & 92.0 (\gain{1.30})  & 91.7 (\gain{0.92}) \\
&  & Style+Term  & 95.6 & 95.7 (\gain{0.01})  & 95.8 (\gain{0.11}) \\
\midrule

\multirow{4}{*}{Qwen3-32B}
& \multirow{4}{*}{Step-by-step}
& Overall     & 68.5 & 66.3 (\gain{-2.22}) & 71.5 (\gain{2.98}) \\
&  & Accuracy    & 88.4 & 88.1 (\gain{-0.34}) & 89.2 (\gain{0.74}) \\
&  & Fluency     & 83.2 & 81.5 (\gain{-1.64}) & 84.4 (\gain{1.25}) \\
&  & Style+Term  & 92.2 & 92.0 (\gain{-0.21}) & 93.0 (\gain{0.73}) \\
\midrule

\multirow{4}{*}{Qwen2.5-72B}
& \multirow{4}{*}{Step-by-step}
& Overall     & 76.5 & 78.1 (\gain{1.61})  & 77.8 (\gain{1.31}) \\
&  & Accuracy    & 92.3 & 91.8 (\gain{-0.53}) & 91.8 (\gain{-0.53}) \\
&  & Fluency     & 86.1 & 87.6 (\gain{1.52})  & 87.4 (\gain{1.29}) \\
&  & Style+Term  & 93.3 & 93.7 (\gain{0.42})  & 94.2 (\gain{0.97}) \\
\midrule

\multirow{4}{*}{Qwen2.5-32B}
& \multirow{4}{*}{Step-by-step}
& Overall     & 64.9 & 65.2 (\gain{0.22})  & 69.2 (\gain{4.30}) \\
&  & Accuracy    & 89.8 & 89.9 (\gain{0.05})  & 90.7 (\gain{0.83}) \\
&  & Fluency     & 79.6 & 80.5 (\gain{0.90})  & 82.4 (\gain{2.83}) \\
&  & Style+Term  & 90.5 & 90.9 (\gain{0.34})  & 91.0 (\gain{0.49}) \\
\midrule

\multirow{4}{*}{Qwen2.5-14B}
& \multirow{4}{*}{Step-by-step}
& Overall     & 61.2 & 64.5 (\gain{3.31})  & 64.6 (\gain{3.39}) \\
&  & Accuracy    & 85.0 & 84.8 (\gain{-0.15}) & 85.5 (\gain{0.52}) \\
&  & Fluency     & 82.9 & 83.6 (\gain{0.71})  & 83.6 (\gain{0.67}) \\
&  & Style+Term  & 92.4 & 92.5 (\gain{0.10})  & 92.6 (\gain{0.23}) \\

\bottomrule
\end{tabular}%
}

\caption{Detailed performance of \textbf{Step-by-step translation}. We report MQM-FSP scores (higher is better) for the Overall, Accuracy, Fluency, and Style+Term dimensions. Following~\citet{briakou2024translating}, we use a 4-step translation pipeline: step1 (\textit{research}), step2 (\textit{drafting}), step3 (\textit{refinement}), and step4 (\textit{proofread}). Note that overall is not the average of dimension scores (see~\ref{appendix:automatic_eval} for details). Deltas are computed from the unrounded scores before rounding for display, so they may not exactly match the difference between the rounded values shown in the table.}
\label{tab:appendix:stepbystep}
\end{table*}

\begin{table*}[t]
\centering
\scriptsize
\setlength{\tabcolsep}{3.0pt}
\renewcommand{\arraystretch}{1.05}

\resizebox{\textwidth}{!}{%
\begin{tabular}{l l l c c c c c}
\toprule
Model & Setting & MQM Dim. & Initial & step1 & step2 & step3 & step4 \\
\midrule

\multirow{24}{*}{DeepSeek-V3-671B}
& \multirow{4}{*}{General}
& Overall     & 85.7 & 87.3 (\gain{1.6})  & 89.7 (\gain{4.0}) & 89.8 (\gain{4.1}) & 89.3 (\gain{3.6})  \\
&            & Accuracy    & 95.7 & 95.5 (\gain{-0.2}) & 95.9 (\gain{0.2}) & 95.8 (\gain{0.1}) & 95.5 (\gain{-0.2}) \\
&            & Fluency     & 93.4 & 94.2 (\gain{0.8}) & 96.4 (\gain{3.0}) & 96.0 (\gain{2.6}) & 96.0 (\gain{2.6}) \\
&            & Style+Term  & 96.6 & 97.6 (\gain{1.0}) & 97.6 (\gain{0.9}) & 98.0 (\gain{1.3}) & 97.8 (\gain{1.2}) \\
\cmidrule(lr){2-8}

& \multirow{4}{*}{Monolingual}
& Overall     & 85.7 & 85.0 (\gain{-0.7})  & 84.5 (\gain{-1.2}) & 85.3 (\gain{-0.3}) & 84.8 (\gain{-0.8})  \\
&            & Accuracy    & 95.7 & 93.0 (\gain{-2.7}) & 92.1 (\gain{-3.6}) & 92.7 (\gain{-3.0}) & 91.9 (\gain{-3.8}) \\
&            & Fluency     & 93.4 & 94.9 (\gain{1.5}) & 94.7 (\gain{1.3}) & 94.8 (\gain{1.4}) & 95.4 (\gain{2.0}) \\
&            & Style+Term  & 96.6 & 97.3 (\gain{0.6}) & 97.8 (\gain{1.2}) & 98.0 (\gain{1.3}) & 97.5 (\gain{0.9}) \\
\cmidrule(lr){2-8}

& \multirow{4}{*}{Eval-Refine}
& Overall     & 85.7 & 88.4 (\gain{2.8})  & 89.7 (\gain{4.0}) & 89.1 (\gain{3.4}) & 89.2 (\gain{3.5})  \\
&            & Accuracy    & 95.7 & 96.0 (\gain{0.3}) & 96.4 (\gain{0.7}) & 96.0 (\gain{0.3}) & 96.0 (\gain{0.3}) \\
&            & Fluency     & 93.4 & 95.1 (\gain{1.7}) & 95.6 (\gain{2.2}) & 95.4 (\gain{2.0}) & 95.9 (\gain{2.5}) \\
&            & Style+Term  & 96.6 & 97.4 (\gain{0.8}) & 97.7 (\gain{1.1}) & 97.8 (\gain{1.1}) & 97.4 (\gain{0.7}) \\
\cmidrule(lr){2-8}

& \multirow{4}{*}{ErrorSpec-Accuracy}
& Overall     & 85.7 & 87.3 (\gain{1.6})  & 88.2 (\gain{2.5}) & 88.0 (\gain{2.3}) & 88.0 (\gain{2.3})  \\
&            & Accuracy    & 95.7 & 95.9 (\gain{0.2}) & 96.6 (\gain{0.9}) & 96.5 (\gain{0.8}) & 96.7 (\gain{1.0}) \\
&            & Fluency     & 93.4 & 94.5 (\gain{1.1}) & 94.8 (\gain{1.4}) & 94.7 (\gain{1.3}) & 94.7 (\gain{1.3}) \\
&            & Style+Term  & 96.6 & 96.9 (\gain{0.3}) & 97.0 (\gain{0.4}) & 96.8 (\gain{0.2}) & 96.6 (\gain{0.0}) \\
\cmidrule(lr){2-8}

& \multirow{4}{*}{ErrorSpec-Fluency}
& Overall     & 85.7 & 85.5 (\gain{-0.2})  & 87.5 (\gain{1.9}) & 87.3 (\gain{1.7}) & 87.7 (\gain{2.0})  \\
&             & Accuracy    & 95.7 & 94.8 (\gain{-0.9}) & 96.1 (\gain{0.4}) & 95.4 (\gain{-0.3}) & 95.9 (\gain{0.2}) \\
&            & Fluency     & 93.4 & 94.2 (\gain{0.8}) & 94.8 (\gain{1.4}) & 94.9 (\gain{1.5}) & 95.0 (\gain{1.6}) \\
&            & Style+Term  & 96.6 & 96.6 (\gain{0.0}) & 96.7 (\gain{0.1}) & 97.1 (\gain{0.5}) & 96.5 (\gain{0.3}) \\
\midrule

\multirow{24}{*}{Qwen3-235B}
& \multirow{4}{*}{General}
& Overall     & 83.0 & 84.0 (\gain{1.0})  & 84.5 (\gain{1.5}) & 85.2 (\gain{2.2}) & 85.6 (\gain{2.6})  \\
&             & Accuracy    & 95.1 & 94.0 (\gain{-1.1}) & 94.1 (\gain{-1.0})	 & 94.7 (\gain{-0.4}) & 94.7 (\gain{-0.4})	\\
&            & Fluency     & 91.1 & 92.7 (\gain{1.7})	& 93.1 (\gain{2.0})	& 93.2 (\gain{2.1}) & 93.7 (\gain{2.6})	 \\
&            & Style+Term  & 96.8 &	97.3 (\gain{0.4})	& 97.5 (\gain{0.6})	& 97.4 (\gain{0.5}) & 97.3 (\gain{0.4})	 \\
\cmidrule(lr){2-8}

& \multirow{4}{*}{Monolingual}
& Overall     & 83.0 & 83.9 (\gain{0.9})  & 83.6 (\gain{0.6}) & 84.6 (\gain{1.6}) & 84.9 (\gain{1.9})  \\
&            & Accuracy    & 95.1 & 92.9 (\gain{-2.2}) & 92.0 (\gain{-3.1}) & 92.2 (\gain{-2.9}) & 92.2 (\gain{-2.9}) \\
&            & Fluency     & 91.1 & 93.7 (\gain{2.6}) & 93.8 (\gain{2.7}) & 94.8 (\gain{3.7}) & 94.9 (\gain{3.9}) \\
&            & Style+Term  & 96.8 & 97.4 (\gain{0.6}) & 98.0 (\gain{1.2}) & 97.6 (\gain{0.8}) & 97.9 (\gain{1.0}) \\
\cmidrule(lr){2-8}

& \multirow{4}{*}{Eval-Refine}
& Overall     & 83.0 & 82.4 (\gain{-0.6})  & 85.8 (\gain{2.8}) & 84.8 (\gain{1.7}) & 86.5 (\gain{3.5})  \\
&             & Accuracy    & 95.1 & 95.0 (\gain{0.0}) & 95.8 (\gain{0.7})	 & 95.2 (\gain{0.1}) & 95.7 (\gain{0.6})	\\
&            & Fluency     & 91.1 & 91.4 (\gain{0.3})	& 93.5 (\gain{2.4})	& 92.8 (\gain{1.7}) & 94.3 (\gain{3.2})	 \\
&            & Style+Term  & 96.8 &	96.0 (\gain{-0.8})	& 96.5 (\gain{-0.3})	& 96.8 (\gain{0.0}) & 96.6 (\gain{-0.2})	 \\
\cmidrule(lr){2-8}

& \multirow{4}{*}{ErrorSpec-Accuracy}
& Overall     & 83.0 & 83.8 (\gain{0.8})  & 85.4 (\gain{2.4}) & 85.2 (\gain{2.2}) & 85.7 (\gain{2.7})  \\
&             & Accuracy    & 95.1 & 95.9 (\gain{0.9})	& 95.9 (\gain{0.8}) &	95.7 (\gain{0.7})	& 95.7 (\gain{0.7}) \\
&            & Fluency     & 91.1 & 91.7 (\gain{0.6})	& 92.9 (\gain{1.9})	& 93.1 (\gain{2.0})	& 93.4 (\gain{2.3})	 \\
&            & Style+Term  & 96.8 &	96.2 (\gain{-0.6})	& 96.6 (\gain{-0.2})	& 96.5 (\gain{-0.3})	& 96.6 (\gain{-0.2})	\\
\cmidrule(lr){2-8}
& \multirow{4}{*}{ErrorSpec-Fluency}
& Overall     & 83.0 & 83.3 (\gain{0.3})  & 82.9 (\gain{-0.1}) & 84.7 (\gain{1.7}) & 84.3 (\gain{1.3})  \\
&             & Accuracy    & 95.1 &	94.8 (\gain{-0.3})	& 94.4 (\gain{-0.7})	& 95.4 (\gain{0.3})	& 95.1 (\gain{0.0})	 \\
&            & Fluency     & 91.1 &	92.0 (\gain{0.9})	& 91.8 (\gain{0.7})	& 93.0 (\gain{1.9})	& 92.7 (\gain{1.6}) \\
&            & Style+Term  & 96.8 &	96.6 (\gain{-0.2})	& 96.8 (\gain{-0.1})	& 96.3 (\gain{-0.5})	& 96.6 (\gain{-0.2})	 \\

\bottomrule
\end{tabular}%
}

\caption{Doc-MT $\rightarrow$ Seg-Refine performance of DeepSeek-V3-671B and Qwen3-235B under different strategy settings.
Each cell reports the absolute score; steps additionally show the change relative to the Initial output in parentheses. Note that overall is not the average of dimension scores (see~\ref{appendix:automatic_eval}). Deltas are computed from the unrounded scores before rounding for display, so they may not exactly match the difference between the rounded values shown in the table.}
\label{tab:doc-seg_all_1}
\end{table*}

\begin{table*}[t]
\centering
\scriptsize
\setlength{\tabcolsep}{3.0pt}
\renewcommand{\arraystretch}{1.05}

\resizebox{\textwidth}{!}{%
\begin{tabular}{l l l c c c c c}
\toprule
Model & Setting & MQM Dim. & Initial & step1 & step2 & step3 & step4 \\
\midrule

\multirow{24}{*}{GPT-OSS-120B}
& \multirow{4}{*}{General}
& Overall     & 74.2 & 78.6 (\gain{4.4})  & 79.8 (\gain{5.6}) & 79.5 (\gain{5.3}) & 80.0 (\gain{5.8})  \\
&             & Accuracy    & 91.3 & 93.8 (\gain{2.5})  & 94.1 (\gain{2.8})	& 94.1 (\gain{2.8}) & 94.2 (\gain{2.8})	\\
&            & Fluency     & 88.3 & 89.3 (\gain{1.0})	& 90.0 (\gain{1.8})	& 89.6 (\gain{1.3}) & 90.0 (\gain{1.7})	 \\
&            & Style+Term  & 94.9 &	96.1 (\gain{1.2})	& 96.0 (\gain{1.1})	& 95.9 (\gain{1.0}) & 96.1 (\gain{1.2})	 \\
\cmidrule(lr){2-8}

& \multirow{4}{*}{Monolingual}
& Overall     & 74.2 & 71.8 (\gain{-2.4})  & 73.0 (\gain{-1.2}) & 71.1 (\gain{-3.1}) & 70.2 (\gain{-4.0})  \\
&            & Accuracy    & 91.3 & 86.1 (\gain{-5.2}) & 86.2 (\gain{-5.1}) & 84.1 (\gain{-7.3}) & 83.2 (\gain{-8.1}) \\
&            & Fluency     & 88.3 & 90.1 (\gain{1.8}) & 90.5 (\gain{2.3}) & 91.0 (\gain{2.7}) & 91.2 (\gain{2.9}) \\
&            & Style+Term  & 94.9 & 95.7 (\gain{0.8}) & 96.6 (\gain{1.7}) & 96.2 (\gain{1.3}) & 95.9 (\gain{1.0}) \\
\cmidrule(lr){2-8}

& \multirow{4}{*}{Eval-Refine}
& Overall     & 74.2 & 76.3 (\gain{2.1})  & 76.6 (\gain{2.4}) & 76.9 (\gain{2.7}) & 76.8 (\gain{2.6})  \\
&            & Accuracy    & 91.3 & 93.9 (\gain{2.6}) & 94.0 (\gain{2.7}) & 94.2 (\gain{2.9}) & 93.8 (\gain{2.4}) \\
&            & Fluency     & 88.3 & 87.4 (\gain{-0.9}) & 87.7 (\gain{-0.6}) & 88.1 (\gain{-0.2}) & 87.8 (\gain{-0.4}) \\
&            & Style+Term  & 94.9 & 95.1 (\gain{0.3}) & 95.5 (\gain{0.6}) & 95.2 (\gain{0.3}) & 95.5 (\gain{0.6}) \\
\cmidrule(lr){2-8}

& \multirow{4}{*}{ErrorSpec-Accuracy}
& Overall     & 74.2 & 75.4 (\gain{1.4})  & 77.0 (\gain{2.8}) & 77.1 (\gain{2.9}) & 77.7 (\gain{3.5})  \\
&             & Accuracy    & 91.3 & 93.6 (\gain{2.3})	& 93.9 (\gain{2.6}) &	94.4 (\gain{3.1})	& 94.8 (\gain{3.4}) \\
&            & Fluency     & 88.3 & 87.1 (\gain{-1.1})	& 88.2 (\gain{-0.1})	& 88.2 (\gain{-0.1})	& 88.9 (\gain{0.6})	 \\
&            & Style+Term  & 94.9 &	94.9 (\gain{0.0})	& 95.2 (\gain{0.3})	& 94.6 (\gain{-0.3})	& 94.4 (\gain{-0.5})	\\
\cmidrule(lr){2-8}
& \multirow{4}{*}{ErrorSpec-Fluency}
& Overall     & 74.2 & 76.0 (\gain{1.8})  & 76.2 (\gain{2.0}) & 76.7 (\gain{2.5}) & 77.2 (\gain{3.0})  \\
&             & Accuracy    & 91.3 &	92.7 (\gain{1.4})	& 92.7 (\gain{1.4})	& 92.1 (\gain{0.8})	& 92.5 (\gain{1.2})	 \\
&            & Fluency     & 88.3 &	87.9 (\gain{-0.4})	& 87.8 (\gain{-0.4})	& 88.8 (\gain{0.5})	& 88.4 (\gain{0.1}) \\
&            & Style+Term  & 94.9 &	95.5 (\gain{0.6})	& 95.7 (\gain{0.8})	& 95.9 (\gain{1.0})	& 96.4 (\gain{1.5})	 \\
\midrule

\multirow{24}{*}{Qwen3-32B}
& \multirow{4}{*}{General}
& Overall     & 70.5 & 73.2 (\gain{2.6})  & 73.8 (\gain{3.3}) & 74.0 (\gain{3.4}) & 74.0 (\gain{3.4})  \\
&            & Accuracy    & 91.2 & 91.4 (\gain{0.2})  & 90.9 (\gain{-0.4})	& 90.6 (\gain{-0.7}) & 90.6 (\gain{-0.6})	\\
&            & Fluency     & 86.0 & 88.0 (\gain{1.9})	& 88.4 (\gain{2.4})	& 88.9 (\gain{2.9}) & 88.4 (\gain{2.4})	 \\
&            & Style+Term  & 93.3 &	93.9 (\gain{0.6})	& 94.7 (\gain{1.4})	& 94.6 (\gain{1.3}) & 95.1 (\gain{1.8})	 \\
\cmidrule(lr){2-8}

& \multirow{4}{*}{Monolingual}
& Overall     & 70.5 & 69.8 (\gain{-0.7})  & 68.5 (\gain{-2.0}) & 68.2 (\gain{-2.3}) & 68.7 (\gain{-1.8})  \\
&            & Accuracy    & 91.2 & 87.7 (\gain{-3.5})  & 86.2 (\gain{-5.1})	& 85.2 (\gain{-6.0}) & 85.0 (\gain{-6.3})	\\
&            & Fluency     & 86.0 & 87.9 (\gain{1.8})	& 88.2 (\gain{2.1})	& 88.8 (\gain{2.8}) & 88.3 (\gain{2.3})	 \\
&            & Style+Term  & 93.3 &	94.3(\gain{1.0})	& 94.2 (\gain{0.9})	& 94.5 (\gain{1.2}) & 95.6 (\gain{2.4})	 \\
\cmidrule(lr){2-8}

& \multirow{4}{*}{Eval-Refine}
& Overall     & 70.5 &73.0 (\gain{2.5})  & 76.4 (\gain{5.8}) & 75.6 (\gain{5.1}) & 77.4 (\gain{6.9})  \\
&            & Accuracy    & 91.2 & 92.4 (\gain{1.1})  & 92.1 (\gain{0.8})	& 92.4 (\gain{1.1}) & 92.4 (\gain{1.1})	\\
&            & Fluency     & 86.0 & 86.4 (\gain{0.3})	& 89.3 (\gain{3.3})	& 88.3 (\gain{2.3}) & 89.3 (\gain{3.3})	 \\
&            & Style+Term  & 93.3 &	94.5(\gain{1.9})	& 95.2 (\gain{1.9})	& 95.2 (\gain{1.9}) & 95.9 (\gain{2.6})	 \\
\cmidrule(lr){2-8}

& \multirow{4}{*}{ErrorSpec-Accuracy}
& Overall     & 70.5 & 74.9 (\gain{4.4})  & 74.9 (\gain{4.3}) & 71.8 (\gain{1.3}) & 73.4 (\gain{2.9})  \\
&            & Accuracy    & 91.2 & 92.7 (\gain{1.4})	& 92.5 (\gain{1.2}) &	92.7 (\gain{1.4})	& 92.9 (\gain{1.6}) \\
&            & Fluency     & 86.0 & 87.7 (\gain{1.7})	& 86.9 (\gain{0.9})	& 85.0 (\gain{-1.0})	& 86.3 (\gain{0.3})	 \\
&            & Style+Term  & 93.3 &	94.6 (\gain{1.3})	& 95.6 (\gain{2.3})	& 94.2 (\gain{1.0})	& 94.3 (\gain{1.0})	\\
\cmidrule(lr){2-8}
& \multirow{4}{*}{ErrorSpec-Fluency}
& Overall     & 70.5 & 70.3 (\gain{-0.3})  & 71.2 (\gain{0.7}) & 71.9 (\gain{1.4}) & 71.7 (\gain{1.2})  \\
&            & Accuracy    & 91.2 &	90.8 (\gain{-0.5})	& 91.3 (\gain{0.1})	& 90.4 (\gain{-0.9})	& 90.5 (\gain{-0.7})	 \\
&            & Fluency     & 86.0 &	86.7 (\gain{0.7})	& 86.7 (\gain{0.8})	& 87.9 (\gain{1.9})	& 87.8 (\gain{1.8}) \\
&            & Style+Term  & 93.3 &	93.0 (\gain{-0.3})	& 93.3 (\gain{0.0})	& 93.8 (\gain{0.5})	& 93.8 (\gain{0.5})	 \\

\bottomrule
\end{tabular}%
}

\caption{Doc-MT $\rightarrow$ Seg-Refine performance of GPT-OSS-120B and Qwen3-32B under different strategy settings.
Each cell reports the absolute score; steps additionally show the change relative to the Initial output in parentheses. Note that overall is not the average of dimension scores (see~\ref{appendix:automatic_eval}). Deltas are computed from the unrounded scores before rounding for display, so they may not exactly match the difference between the rounded values shown in the table.}
\label{tab:doc-seg_all_2}
\end{table*}

\begin{table*}[t]
\centering
\scriptsize
\setlength{\tabcolsep}{3.0pt}
\renewcommand{\arraystretch}{1.05}

\resizebox{\textwidth}{!}{%
\begin{tabular}{l l l c c c c c}
\toprule
Model & Setting & MQM Dim. & Initial & step1 & step2 & step3 & step4 \\
\midrule

\multirow{24}{*}{Qwen2.5-14B}
& \multirow{4}{*}{General}
& Overall     & 64.8 & 65.1 (\gain{0.3})  & 66.9 (\gain{2.0}) & 66.1 (\gain{1.3}) & 67.6 (\gain{2.8})  \\
&            & Accuracy    & 88.5 & 87.9 (\gain{-0.7})  & 89.0 (\gain{0.5})	& 88.2 (\gain{-0.4}) & 89.7 (\gain{1.1})	\\
&            & Fluency     & 83.8 & 83.8 (\gain{0.0})	& 84.0 (\gain{0.2})	& 83.9 (\gain{0.1}) & 83.7 (\gain{-0.1})	 \\
&            & Style+Term  & 92.8 &	93.8 (\gain{1.0})	& 94.0 (\gain{1.2})	& 94.3 (\gain{1.5}) & 94.5 (\gain{1.7})	 \\
\cmidrule(lr){2-8}

& \multirow{4}{*}{Monolingual}
& Overall     & 64.8 & 61.2 (\gain{-3.6})  & 59.4 (\gain{-5.4}) & 59.2 (\gain{-5.7}) & 56.4 (\gain{-8.4})  \\
&            & Accuracy    & 88.5 & 82.9 (\gain{-5.6})  & 80.1 (\gain{-8.4})	& 78.9 (\gain{-9.7}) & 76.1 (\gain{-12.4})	\\
&            & Fluency     & 83.8 & 85.4 (\gain{1.6})	& 85.5 (\gain{1.7})	& 86.4 (\gain{2.6}) & 86.2 (\gain{2.3})	 \\
&            & Style+Term  & 92.8 &	93.1 (\gain{0.3})	& 94.0 (\gain{1.3})	& 94.2 (\gain{1.4}) & 94.6 (\gain{1.8})	 \\
\cmidrule(lr){2-8}

& \multirow{4}{*}{Eval-Refine}
& Overall     & 64.8 & 60.0 (\gain{-4.9})  & 62.4 (\gain{-2.4}) & 67.1 (\gain{2.2}) & 67.2 (\gain{2.4})  \\
&            & Accuracy    & 88.5 & 88.1 (\gain{-0.5})  & 89.3 (\gain{0.7})	& 89.8 (\gain{1.3}) & 89.5 (\gain{1.0})	\\
&            & Fluency     & 83.8 & 78.6 (\gain{-5.2})	& 80.2 (\gain{-3.6})	& 83.6 (\gain{-0.2}) & 83.2 (\gain{-0.6})	 \\
&            & Style+Term  & 92.8 &	93.6 (\gain{0.8})	& 93.3 (\gain{0.5})	& 94.1 (\gain{1.3}) & 94.8 (\gain{2.0})	 \\
\cmidrule(lr){2-8}

& \multirow{4}{*}{ErrorSpec-Accuracy}
& Overall     & 64.8 & 64.0 (\gain{-0.9})  & 63.2 (\gain{-1.6}) & 64.6 (\gain{-0.3}) & 65.3 (\gain{0.5})  \\
&            & Accuracy    & 88.5 & 87.7 (\gain{-0.9})	& 87.8 (\gain{-0.7}) &	89.2 (\gain{0.6})	& 89.9 (\gain{1.4}) \\
&            & Fluency     & 83.8 & 82.1 (\gain{-1.7})	& 81.1 (\gain{-2.8})	& 80.8 (\gain{-3.0})	& 81.2 (\gain{-2.6})	 \\
&            & Style+Term  & 92.8 &	94.6 (\gain{1.8})	& 95.0 (\gain{2.2})	& 94.9 (\gain{2.1})	& 94.6 (\gain{1.8})	\\
\cmidrule(lr){2-8}
& \multirow{4}{*}{ErrorSpec-Fluency}
& Overall     & 64.8 & 63.2 (\gain{-1.6})  & 68.2 (\gain{3.4}) & 65.3 (\gain{0.4}) & 66.8 (\gain{2.0})  \\
&            & Accuracy    & 88.5 &	88.7 (\gain{0.2})	& 92.6 (\gain{4.0})	& 92.6 (\gain{4.0})	& 92.4 (\gain{3.8})	 \\
&            & Fluency     & 83.8 &	80.6 (\gain{-3.3})	& 79.9 (\gain{-3.9})	& 76.7 (\gain{-7.1})	& 78.7 (\gain{-5.1}) \\
&            & Style+Term  & 92.8 &	94.0 (\gain{1.5})	& 95.9 (\gain{3.2})	& 96.3 (\gain{3.8})	& 95.8 (\gain{3.1})	 \\
\midrule

\multirow{24}{*}{Qwen2.5-32B}
& \multirow{4}{*}{General}
& Overall     & 64.4 & 70.8 (\gain{6.4})  & 71.4 (\gain{7.0}) & 71.3 (\gain{6.8}) & 70.5 (\gain{6.1})  \\
&            & Accuracy    & 90.2 & 91.0 (\gain{0.8})  & 91.0 (\gain{0.8})	& 91.2 (\gain{0.9}) & 90.2 (\gain{0.0})	\\
&            & Fluency     & 80.6 & 85.0 (\gain{4.4})	& 85.6 (\gain{5.0})	& 85.7 (\gain{5.2}) & 85.7 (\gain{5.1})	 \\
&            & Style+Term  & 93.6 &	95.1 (\gain{1.5})	& 95.0 (\gain{1.4})	& 94.5 (\gain{0.9}) & 94.8 (\gain{1.2})	 \\
\cmidrule(lr){2-8}

& \multirow{4}{*}{Monolingual}
& Overall     & 64.4 & 66.7 (\gain{2.3})  & 65.4 (\gain{1.0}) & 66.1 (\gain{1.6}) & 65.1 (\gain{0.7})  \\
&            & Accuracy    & 90.2 & 86.9 (\gain{-3.4})  & 84.7 (\gain{-5.5})	& 85.6 (\gain{-4.7}) & 82.6 (\gain{-7.7})	\\
&            & Fluency     & 80.6 & 85.8 (\gain{5.2})	& 86.4 (\gain{5.9})	& 86.4 (\gain{5.9}) & 87.5 (\gain{7.0})	 \\
&            & Style+Term  & 93.6 &	94.1 (\gain{0.5})	& 94.3 (\gain{0.6})	& 94.4 (\gain{0.8}) & 95.1 (\gain{1.4})	 \\
\cmidrule(lr){2-8}

& \multirow{4}{*}{Eval-Refine}
& Overall     & 64.4 & 68.3 (\gain{3.9})  & 72.0 (\gain{7.5}) & 72.3 (\gain{7.9}) & 68.4 (\gain{4.0})  \\
&            & Accuracy    & 90.2 & 91.6 (\gain{1.4})   & 92.0 (\gain{1.7})	& 92.3 (\gain{2.1}) & 91.8 (\gain{1.5})	\\
&            & Fluency     & 80.6 & 83.3 (\gain{2.8})	& 85.3 (\gain{4.7})	& 85.4 (\gain{4.8}) & 82.9 (\gain{2.3})	 \\
&            & Style+Term  & 93.6 &	93.6 (\gain{0.0})	& 94.7 (\gain{1.1})	& 94.8 (\gain{1.2}) & 94.2 (\gain{0.6})	 \\
\cmidrule(lr){2-8}

& \multirow{4}{*}{ErrorSpec-Accuracy}
& Overall     & 64.4 & 70.6 (\gain{6.1})  & 71.4 (\gain{7.0}) & 71.6 (\gain{7.1}) & 68.8 (\gain{4.4})  \\
&            & Accuracy    & 90.2 & 92.1 (\gain{1.9})	& 92.4 (\gain{2.2}) &	92.9 (\gain{2.6})	& 92.4 (\gain{2.1}) \\
&            & Fluency     & 80.6 & 83.1 (\gain{2.5})	& 82.8 (\gain{2.2})	& 82.9 (\gain{2.3})	& 80.9 (\gain{0.4})	 \\
&            & Style+Term  & 93.6 &	95.7 (\gain{2.0})	& 96.3 (\gain{2.7})	& 96.2 (\gain{2.5})	& 95.9 (\gain{2.2})	\\
\cmidrule(lr){2-8}

& \multirow{4}{*}{ErrorSpec-Fluency}
& Overall     & 64.4 & 73.8 (\gain{9.3})  & 73.1 (\gain{8.7}) & 73.7 (\gain{9.3}) & 72.9 (\gain{8.4})  \\
&             & Accuracy    & 90.2 &	93.1 (\gain{2.9})	& 93.6 (\gain{3.3})	& 93.8 (\gain{3.6})	& 93.7 (\gain{3.5})	 \\
&            & Fluency     & 80.6 &	84.5 (\gain{3.9})	& 83.9 (\gain{3.3})	& 83.9 (\gain{3.4})	& 82.5 (\gain{1.9}) \\
&            & Style+Term  & 93.6 &	96.4 (\gain{2.8})	& 96.4 (\gain{2.7})	& 96.3 (\gain{2.6})	& 97.2 (\gain{3.6})	 \\

\bottomrule
\end{tabular}%
}

\caption{Doc-MT $\rightarrow$ Seg-Refine performance of Qwen-2.5 models under different strategy settings.
Each cell reports the absolute score; steps additionally show the change relative to the Initial output in parentheses. Note that overall is not the average of dimension scores (see~\ref{appendix:automatic_eval}). Deltas are computed from the unrounded scores before rounding for display, so they may not exactly match the difference between the rounded values shown in the table.}
\label{tab:doc-seg_all_3}
\end{table*}

\end{document}